\documentclass[10pt,journal,compsoc]{IEEEtran}
\def\IEEEcompsocdiamondline{}

\usepackage{mathptmx}

\usepackage[nocompress]{cite}
\usepackage{amsmath,amssymb}
\usepackage{graphicx}
\usepackage{booktabs}
\usepackage{multirow}
\usepackage{array}
\usepackage{algpseudocode}
\usepackage{xcolor}
\usepackage{tikz}
\usetikzlibrary{arrows.meta,positioning,fit,backgrounds,shapes.geometric,calc}
\usepackage{pgfplots}
\pgfplotsset{compat=1.17}
\usepackage{url}
\usepackage{placeins}
\usepackage[hidelinks,hypertexnames=false]{hyperref}
\hypersetup{pdftitle={LanG 2.0: Energy-Efficient Agentic AI for Semi-Automated Security Operations at Scale}, pdfauthor={Anes Abdennebi, Nadjia Kara, Laaziz Lahlou, Hakima Ould-Slimane}}
\usepackage{algorithm}

\newcommand{\hookw}{\omega}
\newcommand{\ev}{e}

\tikzset{
  blk/.style={draw, rounded corners=2pt, align=center, font=\footnotesize, minimum height=7mm, inner sep=3pt, fill=white},
  src/.style={blk, fill=gray!8},
  core/.style={blk, fill=blue!6},
  llm/.style={blk, fill=orange!10},
  eco/.style={blk, fill=green!8},
  agent/.style={blk, fill=purple!10, draw=purple!65!black},
  review/.style={blk, fill=purple!5, draw=purple!65!black, dashed},
  guard/.style={blk, fill=violet!8, draw=violet!65!black},
  arr/.style={-{Latex[length=2mm]}, thick},
  darr/.style={-{Latex[length=2mm]}, dashed},
  grp/.style={draw, dashed, rounded corners=3pt, inner sep=5pt},
}

\begin{document}

\title{LanG 2.0: Energy-Efficient Agentic AI for Semi-Automated Security Operations at Scale}

\author{Anes~Abdennebi,
        Nadjia~Kara,
        Laaziz~Lahlou,
        and~Hakima~Ould-Slimane
\IEEEcompsocitemizethanks{
\IEEEcompsocthanksitem A. Abdennebi, N. Kara and L. Lahlou are with the Department of Software Engineering and IT, \'Ecole de technologie sup\'erieure (\'ETS), Montreal, QC, Canada.\protect\\
E-mail: anes.abdennebi.1@ens.etsmtl.ca,\protect\\ \{nadjia.kara, laaziz.lahlou\}@etsmtl.ca
\IEEEcompsocthanksitem H. Ould-Slimane is with the Department of Mathematics and Computer Science, Universit\'e du Qu\'ebec \`a Trois-Rivi\`eres (UQTR), Trois-Rivi\`eres, QC, Canada.\protect\\
E-mail: hakima.ould-slimane@uqtr.ca}}

\IEEEtitleabstractindextext{%
\begin{abstract}
Security operations centers (SOCs) need to turn terabytes of alerts, logs and flows into trusted, affordable decisions to unveil hidden threats. We present LanG 2.0, an agentic system that detects threats, correlates evidence and writes rules, investigating whether small local models can replace frontier models affordably and safely when structural safeguards, not model accuracy, provide the guarantees. It incorporates entity-anchored correlation, human approval of rule changes and a tamper-evident ledger. Tests span eight flow datasets from five networks, real XDR alerts and multistage campaigns. Against its predecessor, campaign correlation precision rises from 18.75\% to 51.81\%, and rule acceptance from at most 60.00\% to 98.33\%. Against Claude Opus 5.5, scoring flows against each network's normal traffic detects attacks on unseen networks at 73.42\% F1 without false positives, versus 58.10\% with 49.00\% false positives, for at least 650 times less energy. Finetuned models statistically match it on alert classification and rule writing for 5.6 to 59 times less energy. 
The work demonstrates LanG's practical scalability, processing 3.33 million records in 6.4 minutes, over 30 million hourly, and its safeguards resisted prompt injection, label poisoning and ledger tampering, so small local models can carry most SOC detection affordably and safely.
\end{abstract}

\begin{IEEEkeywords}
Agentic AI, security operations, network intrusion detection, generalization across networks, threat classification, detection rule generation, alert correlation, green AI.
\end{IEEEkeywords}}

\maketitle
\IEEEdisplaynontitleabstractindextext
\IEEEpeerreviewmaketitle

\IEEEraisesectionheading{\section{Introduction}\label{sec:intro}}

\IEEEPARstart{A}{} modern security operations center (SOC) sits at the end of many pipelines. A security information and event management (SIEM) system aggregates firewall, proxy and authentication logs. An extended detection and response (XDR) platform, such as Microsoft Defender XDR, or an endpoint detection and response (EDR) agent streams process, file and registry telemetry. Network sensors export flows and packet captures. Each product raises its own alerts with its own severity scale, its own timestamps and its own notion of an entity. The analyst who receives these alerts has to decide which of them describe the same intrusion, in what order the adversary acted, and which hosts, accounts and files were touched. This work is still largely manual. The 2024 IBM and Ponemon study reports a mean of 194 days to identify a breach and 64 more days to contain it \cite{ibm2024breach}. In one SOC, only 0.01\% of 14.4 million network alerts were associated with true attacks \cite{yang2024true}.

Research on this problem follows two main lines. Alert correlation systems link alerts from several products through shared entities, alert graphs or attack stages \cite{wu2026alert2vec,freitas2024graphweaver,liu2025musar}. Provenance systems build causal graphs from kernel audit records and trace an attack through process, file and socket dependencies \cite{cheng2024kairos,jiang2025orthrus,aly2025ocrapt}, but they need whole-system auditing on every host, which few enterprises deploy. Correlators mostly stop at grouping, and only a few recent systems explain their output in plain language \cite{aly2025ocrapt}. None of them turns the grouped evidence into reviewed detection rules, and none reports what its own computation costs in energy.

Language models are entering SOC tooling for triage, enrichment and report writing \cite{freitas2024guide,xu2024llmsecurity}. A model call is cheap for one alert and expensive for a million. For a widely used generative model, a year of inference emits far more than its training \cite{chien2023reducing}, and generative inference costs orders of magnitude more energy than models built for one task \cite{luccioni2024power}. A SOC running language models unmetered on every event cannot justify that cost.

Our work presents LanG 2.0 and should be distinguished from our previous work \cite{abdennebi2026lang}, which presents the platform's governance model, agent orchestration, operational controls and unified SOC workbench, mechanisms we do not claim as new here. We evaluate how the platform's cooperating agents turn heterogeneous security data into reviewed classifications, log analysis and candidate detection rules grounded in auditable evidence, at what scale and at what energy cost, and we replace the detection and rule generation models of \cite{abdennebi2026lang}. Those detectors were trained and tested within single datasets, whereas we train and test on eight flow datasets from five networks and three exporters, real Microsoft Defender XDR alerts and multistage attack campaigns, and show that supervised flow models do not carry over to unseen networks. Those rule generators reached 90.83\% and 95.50\% acceptance under the validator of that work, but their rules load in the exact engine version requested for only 57.50\% to 60.00\% of our benchmark incidents, against 98.33\% for LanG 2.0.

Our thesis is that a SOC can hand most of the detection phase to small models it trains and runs locally, provided that structural safeguards, rather than model accuracy, carry the guarantees. By structural safeguards we mean four mechanisms that hold whatever a model outputs: every correlation link must share an entity, analysts approve every change to the SOC, only one agent may deploy rules, and a tamper-evident ledger, anchored outside its database, records every action. We test it with three questions. Where do small local agents match a frontier model, and where must a frontier model, a classical model or an analyst take over (Sections~\ref{sec:evalnet} to \ref{sec:evalft})? Do these safeguards hold when telemetry, labels or the ledger are attacked (Section~\ref{sec:evalsec})? What does each step cost in energy (Section~\ref{sec:effort})?

\begin{enumerate}
\item \textbf{An auditable agentic workflow.} An execution graph coordinates detection, classification, log analysis, correlation and rule generation under analyst checkpoints, with an action ledger, guardrails and energy metering for every model call. Autonomous subsystems run the same agents over millions of records without checkpoints and keep every rule in a bucket until an analyst approves it, and two of them analyzed 3.33 million records in 6.4 minutes on one workstation, over 30 million records an hour (Section~\ref{sec:evalscale}).
\item \textbf{Anchored correlation.} The UICR and the eight hooks come from \cite{abdennebi2026lang}. We identify phase collapse, the failure mode of that hook graph, and show that entity anchoring fixes it, raising precision on multistage campaigns from 18.75\% to 51.81\%, that grouping by address remains stronger on real XDR alerts, and that hub pruning keeps anchoring accurate behind a shared egress address (Section~\ref{sec:evalcorr}).
\item \textbf{Detection across networks and rules validated by engines.} Supervised flow classifiers do not transfer between networks, whereas a baseline of a network's normal traffic does, so LanG scores flows against a baseline kept for each tenant. Its rule generator writes rules in the native syntax of six engine versions that load and replay them, not significantly different from Claude Opus 5.5 (Sections~\ref{sec:evalnet} to \ref{sec:evalrules}).
\item \textbf{Where small local models suffice.} Under a preregistered protocol, finetuned agents of 1.7B to 8B parameters match Claude Opus 5.5 on the attack step of real IDS alerts, classical models match them on two of three tasks, and the frontier model remains necessary to name attack families on a new network until about 50 analyst labels exist (Section~\ref{sec:evalft}).
\item \textbf{Safeguards under attack.} Injections from published corpora placed in telemetry, poisoned analyst labels and tampering with the ledger are measured against the pipeline, with what each defense stops and misses (Section~\ref{sec:evalsec}).
\item \textbf{Measured cost.} Energy is measured for every agent operation and for training, and compared with published frontier estimates (Section~\ref{sec:effort}). Comparisons report confidence intervals and paired tests, and the benchmarks evaluate methods, not production deployment.
\end{enumerate}

\section{Related Work}\label{sec:related}

\subsection{Alert Correlation and Aggregation}

Recent correlators link alerts from several products. Alert2Vec embeds each alert through its subgraph in an alert graph to label its threat level, for network and host devices alike \cite{wu2026alert2vec}. GraphWeaver correlates billions of Microsoft Defender XDR alerts that share entities and compresses the result with a minimum spanning tree \cite{freitas2024graphweaver}. MuSAR maps network alarms and application logs to ATT\&CK stages and chains them into multistep attacks across hosts, with an average recall of 93.48\% \cite{liu2025musar}, and on a public set of 2.66 million alerts, two aggregation tools reduce the filtered alerts by 97.73\% and 98.93\% on average \cite{landauer2024alert}. LanG also links sources through shared entities, but requires an entity witness on every graph edge, so that time or kill chain adjacency can weight a link but never create one, and passes its communities to agents that classify the incident and write rules for analyst review. Unlike GraphWeaver, it was not tested on billions of alerts, and its correlation benchmarks are generated, not collected in a SOC.

\subsection{Provenance and Log Based Investigation}

Provenance systems reconstruct attacks from kernel audit graphs. KAIROS, NODLINK, FLASH, MAGIC and AirTag detect anomalous behavior by learning over graphs or raw log text \cite{cheng2024kairos,li2024nodlink,rehman2024flash,jia2024magic,ding2023airtag}, R-CAID embeds root causes into graph learning \cite{goyal2024rcaid}, ORTHRUS cuts the nodes an analyst must inspect by orders of magnitude \cite{jiang2025orthrus}, and OCR-APT has a language model write attack reports from anomalous subgraphs \cite{aly2025ocrapt}, although mimicry attacks reached a 100\% evasion rate against five such detectors in their main evaluation \cite{goyal2023sometimes}. Provenance gives a causal precision that LanG, which never observes process lineage, lacks, at the cost of kernel auditing on every host. Language models also detect anomalies in system logs, finetuned as LogLLM (average F1 score of 0.959 on four datasets \cite{guan2026logllm}) or prompted as LogPrompt (at most 0.450 \cite{liu2024interpretable}), multihost log datasets support such evaluations \cite{landauer2023maintainable}, and Carbon Filter clusters the command lines behind EDR alerts, filtering about 82\% of false alerts at more than 5{,}000 alerts per second \cite{inam2025carbon}. Each method analyzes one stream, whereas LanG correlates the alerts, logs and flows that SIEM and XDR products export and calls a model only on bounded sets of events, at tens of seconds per unit where hashing handles thousands of alerts per second.

\subsection{Network Intrusion Detection by Machine Learning}

Flow-based detectors evaluated on public benchmarks \cite{moustafa2015unsw,moustafa2021toniot,koroniotis2019botiot,sharafaldin2018cicids} often report near perfect scores \cite{layeghy2023crossdomain,apruzzese2023sok}, yet models trained on one dataset generalize poorly to another \cite{layeghy2023crossdomain}, partly because addresses, ports, timestamps and TTL values let them learn the dataset rather than the attack \cite{goldschmidt2025network}. Unsupervised detectors learn normal traffic instead, from autoencoder ensembles such as Kitsune \cite{mirsky2018kitsune} to isolation forests \cite{liu2008isolation} and outlier scores built from empirical distributions such as ECOD \cite{li2023ecod}, and OWAD detects, explains and adapts to shifts of that normality \cite{han2023anomaly}, but they do not name the attack. GPT-4 classifies NetFlow records only slightly better than random guessing \cite{houssel2024explainable}, and flow classifiers can be backdoored by poisoning as little as 0.1\% of their training data \cite{severi2023poisoning}. LanG combines a baseline of each tenant's normal traffic, which decides what is abnormal, with a classifier trained on several networks, which names the attack, and it confirms on four NetFlow datasets that supervised models do not transfer. Unlike OWAD, it refreshes a drifting baseline instead of adapting it continuously, and its retraining on analyst labels is exposed to the poisoning that Severi \textit{et al.} demonstrate, measured in Section~\ref{sec:evalsec}.

\subsection{Language Models in Security Operations}

Language models now parse logs \cite{jiang2024lilac} and assist incident response \cite{freitas2024guide,xu2024llmsecurity}. On ExCyTIn-Bench, the best agent investigating incidents reaches a mean reward of only 0.606 \cite{wu2026excytin}, and in a study with 18 analysts and 50 incidents, summaries written by a model alone omitted critical details in 35\% of cases and contained factual errors in 42\% \cite{kramer2025integrating}. Hammar \textit{et al.} plan incident responses with a finetuned 14B model, retrieval and lookahead, with a bounded probability of hallucination and up to 22\% shorter recovery times than frontier models \cite{hammar2026incident}. For detection rules, SIGMERGE turns commands from threat reports into Sigma rules with components built on Qwen3-1.7B and detected 8 of 10 attacks taken from threat reports \cite{cai2026sigma}, and NeuroYara ranks byte sequences with neural networks to assemble YARA rules \cite{mansour2025neuroyara}. Each of these systems addresses one task, whereas LanG chains them all, accepts a rule only when the engine of the requested version loads it and, in line with Kramer \textit{et al.}, keeps an analyst at every step that changes the SOC. Like Hammar \textit{et al.}, it finds a small local model close to a frontier model on narrow tasks. Unlike SIGMERGE, it does not extract rules from threat reports, and its YARA rules trail those of the frontier model. Finally, models that read text an attacker can influence inherit a new attack surface, since instructions planted in retrieved data can steer them \cite{greshake2023indirect}, and no existing defense stops such injection reliably \cite{liu2024formalizing}. BIPIA found every tested model vulnerable to instructions hidden in external content \cite{yi2025bipia}, and defenses mark untrusted text \cite{hines2024spotlighting} or classify it with a small detector such as Prompt Guard~2 \cite{chennabasappa2025llamafirewall}. LanG adds structural limits to such filters and tests both against BIPIA attacks in its telemetry.

\subsection{Energy and Carbon Accounting for AI}

Inference now dominates the footprint of widely used generative models \cite{chien2023reducing}. Multipurpose generative models cost orders of magnitude more per inference than models built for one task \cite{luccioni2024power}. Measuring inference energy \cite{chung2025mlenergy} is error prone: one carbon calculator reports 400\% more energy than another on the same job, with CodeCarbon closest to a wattmeter \cite{bouza2023how,codecarbon}, estimates from FLOPs understate measured LLM energy \cite{fernandez2025energy}, estimates from TDP overstate it by up to 4.1 times \cite{chung2025mlenergy}, and the power sensor of A100 and H100 GPUs samples only 25\% of the runtime \cite{yang2024accurate}. These works measure models, not security workflows. LanG attributes energy to each agent operation of a SOC, separates the marginal energy of a call from the idle draw, and enforces budgets for each tenant, within the limits above.

\begin{table*}[t]
\centering
\caption{Comparison of LanG with representative correlation, investigation, response, rule generation and accounting approaches.}
\label{tab:related}
\scriptsize
\renewcommand{\arraystretch}{0.92}
\setlength{\tabcolsep}{3.4pt}
\begin{tabular*}{\textwidth}{@{\extracolsep{\fill}}l l c c c c c l@{}}
\toprule
Work & Input data & Multisource & Output & Explanation & Energy report & Budget & Core technique \\
\midrule
Alert2Vec \cite{wu2026alert2vec} & network, host alerts & partial & threat levels & -- & -- & -- & alert subgraph embedding \\
GraphWeaver \cite{freitas2024graphweaver} & XDR alerts & partial & incidents & -- & -- & -- & shared entities, spanning tree \\
MuSAR \cite{liu2025musar} & alarms, application logs & \checkmark & attack chains & -- & -- & -- & ATT\&CK stage linking \\
Carbon Filter \cite{inam2025carbon} & EDR alerts & -- & false alert clusters & -- & -- & -- & similarity hashing \\
KAIROS \cite{cheng2024kairos} & kernel audit & -- & summary graph & -- & -- & -- & temporal GNN \\
ORTHRUS \cite{jiang2025orthrus} & kernel audit & -- & attack paths & -- & -- & -- & GNN, dependency analysis \\
OCR-APT \cite{aly2025ocrapt} & kernel audit & -- & attack reports & \checkmark & -- & -- & subgraph anomaly, LLM \\
Hammar \textit{et al.} \cite{hammar2026incident} & incident logs & -- & response plans & -- & -- & -- & finetuned LLM, lookahead \\
SIGMERGE \cite{cai2026sigma} & threat reports & -- & Sigma rules & -- & -- & -- & small LLM pipeline \\
CodeCarbon \cite{codecarbon} & any program & n/a & n/a & n/a & process level & -- & power sampling \\
\midrule
\textbf{LanG (ours)} & SIEM, XDR, logs, flows & \checkmark & communities, rules & \checkmark & per operation & \checkmark & \shortstack[l]{anchored hooks, site baselines,\\summaries assisted by LLMs} \\
\bottomrule
\end{tabular*}
\end{table*}

Table~\ref{tab:related} summarizes the differences. We avoid a priority claim, which would require an exhaustive survey of SIEM, XDR, provenance and sustainable computing systems.

\section{Problem Statement and Threat Model}\label{sec:model}

Let $\mathcal{S}=\{S_1,\dots,S_L\}$ be a set of log sources, such as a SIEM export, an XDR alert feed or a flow capture, each emitting records in its own format, which a normalizer maps to events
\begin{equation}\label{eq:event}
\ev = (t, s, \mathit{ip}_s, \mathit{ip}_d, p_s, p_d, \pi, h, u, \rho, \rho_p, f, c, I, A, \tau, \sigma),
\end{equation}
where $t$ is a timestamp, $s$ the source, the next five fields the network tuple with protocol $\pi$, $h$ the host, $u$ the account, $\rho$ and $\rho_p$ the process and its parent, $f$ a file path, $c$ a command line, $I$ the set of indicators of compromise, $A$ the set of ATT\&CK techniques, $\tau$ the set of tactics and $\sigma$ the severity. Any field may be empty. Let $E$ be the set of events collected over an investigation window, and Table~\ref{tab:notation} groups the notation of the paper.

\begin{table}[t]
\centering
\caption{Notation, grouped by the section that defines each symbol.}
\label{tab:notation}
\scriptsize
\renewcommand{\arraystretch}{0.9}
\setlength{\tabcolsep}{2.2pt}
\begin{tabular*}{\columnwidth}{@{\extracolsep{\fill}}l l l l@{}}
\toprule
Symbol & Meaning & Symbol & Meaning \\
\midrule
\multicolumn{4}{@{}l}{\textit{Events and correlation (Sections~\ref{sec:model} to \ref{sec:corr})}} \\
$\ev$, $E$ & event (Eq.~\eqref{eq:event}), set & $C$ & attack campaign \\
$\rho$, $\rho_p$ & process, parent process & $\tau$, $\sigma$ & ATT\&CK tactics, severity \\
$I$, $A$ & indicators, ATT\&CK techniques & $c$ & command line \\
$q$ & event confidence (Eq.~\eqref{eq:rel}) & $E_N$ & primary set of $N$ events \\
$h_k$ & hook $k$ (Table~\ref{tab:hooks}) & $\mathcal{H}_E$, $\mathcal{H}_C$ & entity, context hooks \\
$w$, $\gamma$ & weight, damping (Eq.~\eqref{eq:composite}) & $\theta$ & edge threshold (Eq.~\eqref{eq:anchor}) \\
\multicolumn{4}{@{}l}{\textit{Flow detection (Section~\ref{sec:models})}} \\
$\mathcal{B}$, $\mathcal{B}_{\mathrm{ref}}$ & tenant baseline, reference part & $\mathbf{r}(x)$, $n_f$ & ranks of flow $x$, features \\
$a(x)$, $K$ & score, neighbors (Eq.~\eqref{eq:knn}) & $\delta$, $\alpha$ & alert threshold, target FPR \\
\multicolumn{4}{@{}l}{\textit{Execution, ledger and energy (Sections~\ref{sec:agentic} to \ref{sec:carbon})}} \\
$\lambda_k$, $H$ & ledger record, its hash & $N_u$, $N_w$ & work units, workers \\
$E_{\mathrm{tot}}$, $E_{\mathrm{marg}}$ & energies (Eq.~\eqref{eq:energy}) & $P$, $P_{\mathrm{idle}}$ & power, idle power \\
$T$, $u$ & call duration, CPU utilization & $\kappa_g$ & intensity of region $g$ (Eq.~\eqref{eq:co2}) \\
\multicolumn{4}{@{}l}{\textit{Evaluation (Section~\ref{sec:eval})}} \\
$M$, $\phi$ & campaigns, events with address & $\eta$ & campaigns behind one egress \\
$n$ & labels of a new network & $\epsilon$ & share of poisoned labels \\
\bottomrule
\end{tabular*}
\end{table}

An \emph{attack campaign} is a subset $C\subseteq E$ produced by one adversary pursuing one objective. Campaigns may overlap in time, reuse techniques, and move from network to host evidence with no shared field. Benign events are the complement of all campaigns. The \emph{correlation problem} is to find a partition $\hat{\mathcal{C}}$ of $E$ that matches the campaign partition. We evaluate the quality, cost and review volume of that partition, not attack scenario reconstruction.

\textbf{Threat model.} The adversary can compromise hosts and accounts and generate arbitrary benign-looking traffic. We assume that the collection pipeline is trusted, that at least part of the adversary's activity is visible in at least one source, and that timestamps are accurate within the correlation window after normalization. Log tampering and evasion of the upstream detectors are out of scope. The language model runs locally, so evidence stays in the SOC, and its outputs are untrusted suggestions that must be grounded in events. The adversary also controls text that reaches the agents through telemetry, such as request paths, user agents, log messages and alert names, and may plant instructions there to lower a verdict, change a technique, steer a rule or leak data. A fraction of the analyst labels that adapt the flow classifier may be wrong or adversarial, and labels approved by an autonomous policy may be the system's own errors. For the action ledger, we assume an adversary with write access to its database but not to the ledger key or the store of its anchors. Malicious analysts are covered only as sources of poisoned labels. LanG can also call frontier models such as GPT, Claude and Gemini, which take evidence out of the SOC and escape the carbon plane, so we use one, Claude Opus 5.5, only as an external baseline on data without tenant information, with published energy figures.

\section{Methodology}\label{sec:method}

LanG turns the output of many security tools into reviewed decisions through a chain of stages, each shrinking the evidence that the next stage, and finally the analyst, must examine. Records from all tools are merged into incidents (Section~\ref{sec:uicr}), log analysis keeps a bounded primary set of relevant events (Section~\ref{sec:loganalysis}), and the anchored multihook graph groups it into evidence communities (Section~\ref{sec:corr}). Detection and rule generation agents (Section~\ref{sec:models}) work with classification, log analysis and correlation agents in a workflow with analyst checkpoints (Section~\ref{sec:agentic}), which autonomous subsystems run over millions of records without checkpoints, keeping their rules in buckets (Section~\ref{sec:autonomy}), and a carbon plane meters every model call (Section~\ref{sec:carbon}). The analyst thus reviews a few evidence communities and rule buckets instead of every record or one workflow per alert. Fig.~S1 of the supplement shows the data and evidence plane and Fig.~\ref{fig:agentic} the agentic execution plane. Connectors ingest SIEM, XDR, EDR, network capture, log and threat intelligence data, and every store serves one tenant.

\subsection{Correlating Incidents Across Tools}\label{sec:uicr}

Each tool reports its own fragment of an intrusion, such as an authentication burst in the SIEM, a process tree in the EDR or a flow at a network sensor. The first stage, from our previous work \cite{abdennebi2026lang}, joins these fragments into Unified Incident Context Records (UICRs) with exact merge rules, so that later stages reason over incidents rather than over isolated alerts. Two records merge when they share a correlation key, such as a five-tuple or an alert identifier, the fingerprint of an indicator, or an address within the incident window. After each merge, a bounded triage score combines the indicators, the alerts, the furthest kill chain phase, with ATT\&CK tactics \cite{strom2018attack} mapped to the seven Lockheed Martin phases \cite{hutchins2011killchain}, and the confidence of the local detection models. It orders the analyst queue and feeds the relevance ranking of the next stage. Section~S2 of the supplement gives the record, merge predicates, score and incremental algorithm.

\subsection{Efficient Log Analysis}\label{sec:loganalysis}

An incident can hold thousands of log events, more than an analyst can read or a prompt can hold, so LanG reduces them before any model reasons over them and every model call stays bounded. Events are first deduplicated by fingerprint and scored with
\begin{equation}\label{eq:rel}
\begin{split}
\mathrm{rel}(\ev)=\min\big(1,\ &0.3q+\min(0.2,0.1|I|)+\min(0.25,0.15|A|)\\
&+\beta(\sigma)+0.05\,\mathbb{1}_{\mathit{ip}}+0.03\,\mathbb{1}_{\mathit{port}}\\
&+0.05\,\mathbb{1}_{c}+0.02\,\mathbb{1}_{\rho}\big),
\end{split}
\end{equation}
where $q$ is the event confidence and $\beta$ maps severity to $\{0.25,0.2,0.12,0.05,0.01\}$. The top $N$ events (100 by default) form the primary set $E_N$. Among them, events with no technique, no alert name and confidence below 0.7 are sent to the model for semantic tagging, with at most 30 calls per scan. The log analysis agent then analyzes the leading event of a workflow's evidence, never the raw stream, and returns a summary, a risk level, recommended actions and typed indicators in at most 800 output tokens, for the rule planning agent and the analyst. Its cost therefore does not grow with the volume of the logs. For millions of lines, the front stage of the autonomous subsystems (Section~\ref{sec:autonomy}) reduces the input before incidents are even formed. It keeps only the records whose level is severe or whose message carries an attack indicator, and masks variable fields such as addresses, numbers and identifiers, so that lines of one event type share a message pattern. One work unit then stands for all the suspicious lines of a pattern, service and host.

\subsection{Anchored Multihook Event Correlation}\label{sec:corr}

UICR merging is exact and fast, but it only joins records that share a key, an indicator or an address. It cannot link a brute force burst seen by the SIEM to the account change that the EDR reports on a host minutes later. The second stage addresses this gap with a graph of events whose edges combine several weak signals.

\subsubsection{Correlation Hooks}

A hook is a function $h_k:E\times E\rightarrow[0,1]$ that measures one kind of relation between two events. Table~\ref{tab:hooks} lists the eight hooks. The temporal hook decays exponentially inside a window $W$:
\begin{equation}
h_{\mathrm{tmp}}(\ev_i,\ev_j)=\begin{cases}\exp\!\left(-\dfrac{3|t_i-t_j|}{W}\right) & |t_i-t_j|\leq W,\\ 0 & \text{otherwise.}\end{cases}
\end{equation}
The kill chain hook rewards adjacency of phases. With $d=|\mathrm{ord}(\tau_i)-\mathrm{ord}(\tau_j)|$, it returns 0.8 for $d=0$, 1.0 for $d=1$, 0.6 for $d=2$ and $\max(0,0.4-0.1d)$ beyond. The behavioral hook adds evidence from identical processes, parent and child relations, command-line Jaccard similarity above 0.3 and identical file paths.

\begin{table}[t]
\centering
\caption{Correlation hooks. Hooks marked E are entity hooks and may anchor an edge. Hooks marked C are context hooks.}
\label{tab:hooks}
\scriptsize
\begin{tabular*}{\columnwidth}{@{\extracolsep{\fill}}l c p{5.55cm}@{}}
\toprule
Hook & Class & Weight rule \\
\midrule
Temporal & C & $\exp(-3\Delta t/W)$ for $\Delta t\le W$ ($W=300$\,s) \\
IP linkage & E & 1.0 shared address, 0.8 NAT pivot ($\mathit{ip}_s\!=\!\mathit{ip}_d'$), 0.7 same /24 \\
Host cooccurrence & E & 0.85 same host within 120\,s, 0.4 same host otherwise \\
Flow pattern & C & protocol 0.3, destination port 0.4 (0.15 if within 100), source port 0.2, volume ratio above 0.8 adds 0.1 \\
IoC overlap & E & $\min(1,0.3\,|I_i\cap I_j|)$ \\
Kill chain & C & phase distance rule (0.8, 1.0, 0.6, decaying) \\
Behavioral & E & process 0.5, parent and child 0.7, command Jaccard $\times0.5$, file 0.4 \\
User session & E & 0.9 same account \\
\bottomrule
\end{tabular*}
\end{table}

\subsubsection{Composite Weight and Anchoring}

Let $\hookw_{(1)}\geq\hookw_{(2)}\geq\dots$ be the positive hook weights of a pair sorted in decreasing order. LanG combines them with geometric damping so that many weak signals cannot outweigh one strong signal:
\begin{equation}\label{eq:composite}
w(\ev_i,\ev_j)=\min\Big(1,\ \sum_{m\geq1}\hookw_{(m)}\,\gamma^{m-1}\Big),\qquad \gamma=0.7.
\end{equation}
Our first implementation added an edge whenever $w\geq\theta$ with $\theta=0.15$, which fails (Section~\ref{sec:eval}). Two attack events from different campaigns in adjacent phases receive a kill chain weight of 1.0, plus a temporal weight when close in time, so context hooks connect all campaigns through each phase and modularity optimization returns one community per group of phases, an effect we call \emph{phase collapse}. The fix requires an entity witness:
\begin{equation}\label{eq:anchor}
(\ev_i,\ev_j)\in\mathcal{E}\iff w(\ev_i,\ev_j)\geq\theta\ \wedge\ \exists k\in\mathcal{H}_E:\ h_k(\ev_i,\ev_j)>0,
\end{equation}
where $\mathcal{H}_E$ is the set of entity hooks in Table~\ref{tab:hooks}. Context hooks still shape the weight of anchored edges and thus the community structure, but they can no longer create an edge alone.

\subsubsection{Community Detection and Super Nodes}

LanG runs Louvain modularity optimization \cite{blondel2008louvain} on the weighted graph with resolution 1.0 and a fixed seed. When a community exceeds 100 events, the events of each phase beyond the ten most relevant are folded into a super node that keeps the five most relevant events, the number folded and their mean confidence. This bounds the model context, and the analyst can expand the super node. Algorithm~S2 of the supplement summarizes the stage.

\textbf{Complexity.} Hook evaluation is constant time per pair, so graph construction costs $O(|\mathcal{H}|\,N^2)$ and Louvain runs in near-linear time in the number of edges. The top $N$ cut bounds the quadratic term regardless of the raw event volume, and UICR merging, which is linear in the number of incidents per record, runs before it. Section~\ref{sec:eval} measures both.

\subsection{Detection and Rule Generation Agents}\label{sec:models}

\subsubsection{Flow Detection Across Networks}

\textbf{Shared representation.} Flow exporters disagree on units, directions and feature sets. LanG maps every flow, whether a NetFlow v3, CICFlowMeter or NTLFlowLyzer record or a flow assembled from a packet capture, to 24 fields that all of them provide, in NetFlow units: protocol, destination port, packets and bytes per direction, duration, extreme packet lengths, interarrival statistics, throughput, TCP flags and window sizes. It adds bytes per packet in each direction, the packet rate, the ratio of returned to sent bytes and the TCP flag bits. Addresses, source ports, timestamps and TTL values are withheld, since they let a model learn the dataset rather than the attack \cite{goldschmidt2025network}. Labels map to seven families (benign, denial of service, reconnaissance, exploitation, brute force, malware and command and control, and collection), each tied to an ATT\&CK tactic.

\textbf{Site baseline.} Each tenant keeps a baseline $\mathcal{B}$ of at least 1{,}000 benign flows (5{,}000 here) sampled over the capture period. Quantile grids of $\mathcal{B}$ map the $n_f=23$ numeric features of a flow $x$ to percentile ranks $\mathbf{r}(x)\in[0,1]^{n_f}$, and the anomaly score is the mean distance to the $K=5$ nearest reference flows in rank space,
\begin{equation}\label{eq:knn}
a(x)=\frac{1}{K}\sum_{k=1}^{K}\big\lVert\mathbf{r}(x)-\mathbf{r}(b_{(k)})\big\rVert_2 ,\qquad b_{(k)}\in\mathcal{B}_{\mathrm{ref}},
\end{equation}
where $\mathcal{B}_{\mathrm{ref}}$ holds 70\% of the baseline. The threshold $\delta$ is the $(1-\alpha)$ quantile of $a$ over the other 30\%, for a false positive rate $\alpha=0.01$. The features furthest in the baseline's tails explain alerts.

\textbf{Family classifier.} Gradient boosted trees \cite{ke2017lightgbm} name the family from the same ranks, the protocol and the TCP flag bits, trained with class balancing weights on eight datasets from five networks, each ranked against its own baseline. A flow is an attack if $a(x)>\delta$ or if the classifier gives an attack family a probability above 0.9. When the baseline flags a flow that the classifier calls benign, the agent reports an unknown anomaly and its most likely family.

\textbf{Adaptation and drift.} Analyst decisions at the classification checkpoint become labels: an approval confirms the detected family, a modification supplies the analyst's family, and a rejection marks the flow benign. Only decisions of human principals are exported, so approvals by an autonomous policy never become labels. A label audit sends a benign label on a flow that the baseline finds anomalous to a second analyst before use. Once 50 flows of a tenant are labeled, the classifier is retrained on a replay sample of its training data plus the labeled flows, weighted 20 times, for that tenant only. A holdout gate keeps 25\% of the labels aside, stratified by family, and lets the candidate replace the current model only if its benign false positive rate stays below $\max(1.5\alpha,\ \text{current}+0.5)$ points and neither its macro F1 nor its recall on a canary sample of the training networks drops by more than two points. A population stability index above 0.25 on any feature between recent traffic and the baseline flags it for refresh. Since the baseline alone decides what is anomalous, feedback changes how alerts are named and can add confident alerts but never removes one the baseline raises.

\subsubsection{Rule Generation Validated by Engines}

\textbf{Data.} The training set holds 36{,}553 pairs of incidents and rules built from Emerging Threats Open (Snort~2.9 and Suricata~6 and~7 editions), the Snort community rules for Snort~2.9 and Snort~3, Emerging Threats rules converted to Snort~3 with \texttt{snort2lua}, public YARA rules and SigmaHQ. Unlike the data of \cite{abdennebi2026lang}, whose ``Snort'' targets were mostly Suricata rules, each target is written in the native syntax of the engine and version the input names. Each target is grounded, so every string it matches appears in the input as an indicator, and rules needing regular expressions, byte tests or flow bits are excluded. Vendor metadata is removed, identifiers are moved to a local range, every target is loaded by its engine, and the data are split by rule template across engines, so a detection trained for one engine never appears in another engine's test set.

\textbf{Model and serving.} A single model serves all six targets: Qwen3-1.7B \cite{qwen2025qwen3} adapted with LoRA \cite{hu2022lora} of rank 32, trained on the rule tokens for two epochs (2.5\,h on one GPU, a run whose energy was not metered and is bounded in Section~\ref{sec:effort}). Its prompt states the engine and version, threat name, category, application protocol, references and indicators. For serving, the adapter is merged, quantized to 8 bits and run by llama.cpp \cite{llamacpp} through Ollama, on the same metered path as every other local model. An incident with addresses but no content indicator receives a deterministic indicator rule, so the model never invents payloads.

\subsubsection{Finetuned Agents for the Detection Phase}

Three agents of the detection phase can run a small finetuned model instead of a general one. The \emph{alert analyst} reads one IDS or HIDS alert (source, rule, groups, program and log line) and names the attack step it shows, among ten steps and none, with its ATT\&CK technique. The \emph{incident grader} reads a Defender XDR incident, rendered as a header and up to 12 alerts with their category, techniques and entity roles, and grades it as a true positive, a benign positive or a false positive. The \emph{family namer} reads a flow that the site baseline flagged, with its values and units, the service of its port, its decoded flags, the five features furthest in the baseline's tails and the anomaly score, and names its family or calls it a false alarm. Each agent answers in a JSON schema and its prompt never names a data set, network or organization.

Each agent is a LoRA adapter on a Qwen3 base model \cite{qwen2025qwen3}, trained on the completion only. The base grows only when needed: an agent starts at 1.7B parameters and moves to 4B, then to 8B with 4-bit QLoRA \cite{dettmers2023qlora}, only if its validation macro F1 stays below that of the frontier model on the same validation items (the alerts, incidents or flows of the validation split, one prompt each) and the previous step gained at least two points. The adapters are merged, quantized to 8 bits and served through Ollama with greedy decoding, on the metered path of every other model. Training energy is the difference of the GPU's NVML energy counter read just before and after the training loop, plus a CPU share of about 1\%, so it covers the training run alone, and inference is metered for each call.

\subsection{Semi-Autonomous Multiagent Execution}\label{sec:agentic}

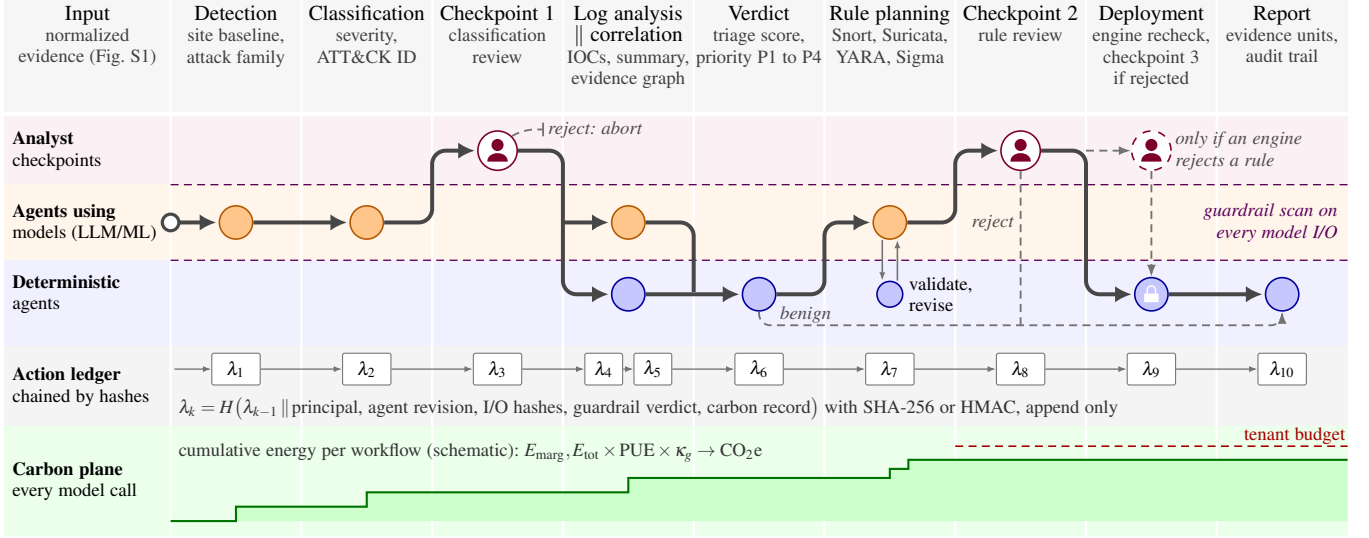
\begin{figure*}[t]
\centering
\begin{tikzpicture}[x=1mm, y=1mm, font=\scriptsize,
  station/.style={circle, draw, line width=0.6pt, minimum size=4.4mm, inner sep=0pt},
  mstation/.style={station, fill=orange!45, draw=orange!70!black},
  dstation/.style={station, fill=blue!22, draw=blue!60!black},
  hstation/.style={station, fill=white, draw=purple!65!black, minimum size=5.2mm},
  ctl/.style={line width=1.5pt, black!72, rounded corners=1.8mm, -{Latex[length=2.2mm, width=2.2mm]}, shorten >=0.3mm},
  alt/.style={line width=0.7pt, black!55, densely dashed, rounded corners=1.5mm, -{Latex[length=1.6mm]}},
  rec/.style={draw=black!55, fill=white, rounded corners=0.8pt, minimum width=6.4mm, minimum height=4.2mm, inner sep=0pt, font=\scriptsize},
  link/.style={-{Latex[length=1.1mm]}, black!55, line width=0.5pt},
  hdr/.style={anchor=north, align=center, inner sep=0pt, text=black!75},
  lanelbl/.style={anchor=west, align=left, inner sep=0pt},
  analyst/.pic={
    \fill[purple!65!black] (0,0.75) circle (0.8);
    \fill[purple!65!black] (-1.35,-1.45) .. controls (-1.35,0.05) and (1.35,0.05) .. (1.35,-1.45) -- cycle;
  },
  padlock/.pic={
    \draw[white, line width=0.55pt] (-0.55,0) -- (-0.55,0.55) arc (180:0:0.55) -- (0.55,0);
    \fill[white, rounded corners=0.2pt] (-0.95,-1.05) rectangle (0.95,0.1);
  },
]
\def\xR{177.7}
\def\yA{54.5} \def\yM{45} \def\yD{35.5} \def\yB{31.4}
\def\yL{25.6}

\fill[gray!7]   (0,59.6) rectangle (\xR,74.6);
\fill[purple!6] (0,50) rectangle (\xR,59);
\fill[orange!8] (0,40) rectangle (\xR,50);
\fill[blue!6]   (0,28.6) rectangle (\xR,40);
\fill[gray!9]   (0,18) rectangle (\xR,28.6);
\fill[green!8]  (0,3.5) rectangle (\xR,18);
\foreach \x in {22,39.3,56.6,73.9,91.2,108.5,125.8,143.1,160.4}
  \draw[white, line width=0.9pt] (\x,3.5) -- (\x,74.6);
\draw[violet!70!black, densely dashed, line width=0.5pt] (22,50) -- (\xR,50) (22,40) -- (\xR,40);

\node[lanelbl] at (1,\yA) {\textbf{Analyst}\\checkpoints};
\node[lanelbl] at (1,\yM) {\textbf{Agents using}\\models (LLM/ML)};
\node[lanelbl] at (1,\yD) {\textbf{Deterministic}\\agents};
\node[lanelbl] at (1,23.3) {\textbf{Action ledger}\\chained by hashes};
\node[lanelbl] at (1,10.75) {\textbf{Carbon plane}\\every model call};

\node[hdr] at (11,73.8)    {{\footnotesize\color{black} Input}\\normalized\\evidence (Fig.~S1)};
\node[hdr] at (30.65,73.8) {{\footnotesize\color{black} Detection}\\site baseline,\\attack family};
\node[hdr] at (47.95,73.8) {{\footnotesize\color{black} Classification}\\severity,\\ATT\&CK ID};
\node[hdr] at (65.25,73.8) {{\footnotesize\color{black} Checkpoint 1}\\classification\\review};
\node[hdr] at (82.55,73.8) {{\footnotesize\color{black} Log analysis}\\{\footnotesize\color{black} $\parallel$ correlation}\\IOCs, summary,\\evidence graph};
\node[hdr] at (99.85,73.8) {{\footnotesize\color{black} Verdict}\\triage score,\\priority P1 to P4};
\node[hdr] at (117.15,73.8){{\footnotesize\color{black} Rule planning}\\Snort, Suricata,\\YARA, Sigma};
\node[hdr] at (134.45,73.8){{\footnotesize\color{black} Checkpoint 2}\\rule review};
\node[hdr] at (151.75,73.8){{\footnotesize\color{black} Deployment}\\engine recheck,\\checkpoint 3\\if rejected};
\node[hdr] at (169.05,73.8){{\footnotesize\color{black} Report}\\evidence units,\\audit trail};

\node[mstation] (det)  at (30.65,\yM) {};
\node[mstation] (cls)  at (47.95,\yM) {};
\node[hstation] (rv1)  at (65.25,\yA) {}; \pic at (rv1) {analyst};
\node[mstation] (log)  at (82.55,\yM) {};
\node[dstation] (cor)  at (82.55,\yD) {};
\node[dstation] (ver)  at (99.85,\yD) {};
\node[mstation] (rul)  at (117.15,\yM) {};
\node[dstation, minimum size=3.4mm] (val) at (117.15,\yD) {};
\node[hstation] (rv2)  at (134.45,\yA) {}; \pic at (rv2) {analyst};
\node[dstation] (dep)  at (151.75,\yD) {}; \pic at (dep) {padlock};
\node[hstation, dashed] (rv3) at (151.75,\yA) {}; \pic at (rv3) {analyst};
\node[dstation] (rep)  at (169.05,\yD) {};

\draw[line width=1.2pt, black!72, fill=white] (22,\yM) circle (1.1mm);
\draw[ctl] (23.1,\yM) -- (det);
\draw[ctl] (det) -- (cls);
\draw[ctl] (cls) -- (56.6,\yM) -- (56.6,\yA) -- (rv1);
\draw[ctl] (rv1) -- (73.9,\yA) -- (73.9,\yD) -- (cor);
\draw[ctl] (73.9,\yM) -- (log);
\draw[ctl, -] (log) -- (91.2,\yM) -- (91.2,\yD);
\draw[ctl] (cor) -- (ver);
\draw[ctl] (ver) -- (108.5,\yD) -- (108.5,\yM) -- (rul);
\draw[ctl] (rul) -- (125.8,\yM) -- (125.8,\yA) -- (rv2);
\draw[ctl] (rv2) -- (143.1,\yA) -- (143.1,\yD) -- (dep);
\draw[alt] (143.1,\yA) -- (rv3);
\draw[alt] (rv3) -- (dep);
\node[anchor=west, inner sep=0.8pt, align=left, text=black!70, font=\scriptsize\itshape] at (154.6,\yA) {only if an engine\\rejects a rule};
\draw[ctl] (dep) -- (rep);

\draw[link, line width=0.6pt] ([xshift=-1mm]rul.south) -- ([xshift=-1mm]val.north |- val.north);
\draw[link, line width=0.6pt] ([xshift=1mm]val.north) -- ([xshift=1mm]rul.south |- rul.south);
\node[anchor=west, inner sep=1pt, align=left] at (119.3,\yD) {validate,\\revise};

\draw[alt, -{Bar[width=1.8mm]}] (rv1.north east) -- (68.6,57.3) -- (71.4,57.3);
\node[anchor=west, inner sep=0.8pt, text=black!70, font=\scriptsize\itshape] at (71.6,57.3) {reject: abort};
\draw[alt] (ver.south) |- (166,\yB) -| (rep.south);
\draw[alt, -] (rv2.south) -- (134.45,\yB);
\node[anchor=south west, inner sep=0.8pt, text=black!70, font=\scriptsize\itshape] at (102.3,\yB) {benign};
\node[anchor=east, inner sep=0.8pt, text=black!70, font=\scriptsize\itshape] at (133.9,\yM) {reject};

\node[violet!70!black, anchor=east, align=right, inner sep=1pt, font=\scriptsize\itshape] at (176.8,\yM) {guardrail scan on\\every model I/O};

\foreach \x/\k in {30.65/1, 47.95/2, 65.25/3, 99.85/6, 117.15/7, 134.45/8, 151.75/9, 169.05/10}
  \node[rec] (r\k) at (\x,\yL) {$\lambda_{\k}$};
\node[rec, minimum width=5mm] (r4) at (79.3,\yL) {$\lambda_4$};
\node[rec, minimum width=5mm] (r5) at (85.8,\yL) {$\lambda_5$};
\draw[link] (22.5,\yL) -- (r1);
\foreach \a/\b in {1/2, 2/3, 3/4, 4/5, 5/6, 6/7, 7/8, 8/9, 9/10}
  \draw[link] (r\a) -- (r\b);
\node[anchor=west, inner sep=0pt, black!80] at (23,20.4)
  {$\lambda_k = H\big(\lambda_{k-1} \,\Vert\, \text{principal, agent revision, I/O hashes, guardrail verdict, carbon record}\big)$ with SHA-256 or HMAC, append only};

\fill[green!20] (22,5.5) -- (30.65,5.5) -- (30.65,7.4) -- (47.95,7.4) -- (47.95,9.3) -- (82.55,9.3) -- (82.55,11.2)
  -- (117.15,11.2) -- (117.15,12.4) -- (119.6,12.4) -- (119.6,13.6) -- (\xR,13.6) -- (\xR,5.5) -- cycle;
\draw[green!45!black, line width=0.8pt] (22,5.5) -- (30.65,5.5) -- (30.65,7.4) -- (47.95,7.4) -- (47.95,9.3) -- (82.55,9.3) -- (82.55,11.2)
  -- (117.15,11.2) -- (117.15,12.4) -- (119.6,12.4) -- (119.6,13.6) -- (\xR,13.6);
\draw[red!65!black, densely dashed, line width=0.6pt] (125.8,15.4) -- (\xR,15.4);
\node[anchor=south east, inner sep=0.6pt, red!65!black] at (\xR,15.4) {tenant budget};
\node[anchor=west, inner sep=0pt, black!80] at (23,14.6)
  {cumulative energy per workflow (schematic): $E_{\mathrm{marg}}, E_{\mathrm{tot}} \times \mathrm{PUE} \times {\kappa_g} \to \mathrm{CO_2e}$};
\end{tikzpicture}
\caption{Agentic execution plane as an execution trace. Columns are agent graph steps, lanes the principal in control. Orange stations call a model, blue ones are deterministic. The analyst reviews the classification and the rules, and decides at the dashed third checkpoint on rules an engine rejects. Only the deployment agent (padlock) writes rules. Model inputs and outputs cross a guardrail (violet), every action is appended to the hash chained ledger, and energy accrues only at model calls.}
\label{fig:agentic}
\end{figure*}

The stages above feed a graph of cooperating agents (Fig.~\ref{fig:agentic}), semi-automated so that a human stays in the loop. At the classification checkpoint an analyst can approve, reject or modify the result, a second checkpoint precedes storing or deploying rules, and a third follows an engine's rejection of an approved rule.

\textbf{Execution order.} The agent graph runs in supersteps whose parallel outputs are merged in a fixed order, so parallelism does not change the outcome. The detection agent scores the flow against the tenant's traffic baseline and names its attack family (Section~\ref{sec:models}), and the classification agent returns a severity and an ATT\&CK technique that is checked against a local copy of the catalog. The workflow then pauses at the first checkpoint, and a rejection ends it. After approval, the log analysis agent and the correlation agent run in the same superstep, the latter supplying the anchored evidence communities of Section~\ref{sec:corr}. A deterministic verdict then assigns a triage score in $[0,100]$ and a priority from P1 to P4. A benign verdict sends the incident directly to the report.

\textbf{Rule planning and deployment.} For a threat verdict, the rule planning agent selects formats from the evidence: Snort~2, Snort~3 and Suricata for network evidence, Sigma for log evidence, and YARA in every case. One author agent per format writes a rule with the finetuned generator of Section~\ref{sec:models}. A deterministic validator checks each rule statically, loads it into the engine and version it targets, then replays it against the captured traffic when available, or against the workflow logs for Sigma. An invalid rule returns to its author for revision, at most twice by default. A consistency check flags disagreements between agents, such as a severity that contradicts the triage score. Rules approved or edited at the second checkpoint are loaded again into their engines, and if an engine rejects one, a third checkpoint lets the analyst deploy, correct or withhold it, since one invalid rule can stop a sensor from loading its ruleset. Only the deployment agent writes to the rule store, and it refuses to write without the required decisions.

\textbf{Identity, permissions and ledger.} Each agent has a stable identity, a version and a declared tool set, and its resolved configuration (version, model binding, prompt template and tools) is hashed at every run into an agent revision, so a change of model or prompt is always attributable. A permission layer checks every tool and model call against the declared tools and records any refusal, and a guardrail sentinel scans every model input and output for prompt injection and data leakage. At intake, every untrusted string of an incident (packet fields, log messages, alert names and indicators) is normalized (Unicode folding, zero-width characters, escapes and whitespace) and scanned on its own. A blocked field is replaced by a placeholder in what the models see and kept as evidence, and the incident is marked as suspected of injection, so it never receives a benign verdict and an autonomous policy abstains on it, while the workflow never halts, since halting would let one string suppress an incident. Prompt Guard~2 \cite{chennabasappa2025llamafirewall} can add a learned layer to this scan, and spotlighting \cite{hines2024spotlighting} can mark untrusted text in prompts. Agents may only write \texttt{alert} rules, so a \texttt{pass} or \texttt{drop} rule is invalid unless an analyst overrides it, indicators a model names are kept only if they occur in the evidence, and internal addresses and secrets never reach external threat intelligence. Every action (node run, model or tool call, guardrail scan, routing decision, checkpoint or analyst decision) is appended to a ledger kept for each tenant. Record $\lambda_k$ stores the principal, the agent revision, hashes of the prompt, input and output, redacted snapshots, token counts and the emission record identifier, and its hash covers this content and the previous record, $\lambda_k=H(\lambda_{k-1}\,\Vert\,\mathrm{content}_k)$, where $H$ is SHA-256, or HMAC-SHA-256 with a ledger key. Database triggers reject updates and deletions, verification recomputes the chain, and since without a key an adversary with write access could rebuild it, keyed hashing is recommended. A chain alone cannot reveal removed last records or a restored backup, so the ledger exports signed anchors (head sequence number, hash and count) to a store outside the database, against which verification detects truncation and rollback. A record that cannot be written goes to an append-only spool and marks the ledger degraded, and critical actions (analyst decisions and rule writes) then fail closed.

\subsection{Scaling with Autonomous Subsystems}\label{sec:autonomy}

One workflow costs tens of seconds of inference and each checkpoint costs analyst attention, so for millions of flows and log lines a day, one workflow per alert would take days of computation and thousands of checkpoint decisions. For such inputs, analysts launch \emph{autonomous subsystems}, choosing the sources and the budget of workflows and workers, possibly several at once, for example one per data source. Each runs the agent graph of Section~\ref{sec:agentic} with the same agents, tools, guardrails, ledger and carbon plane, and differs from an interactive workflow in three ways (Fig.~S5, Algorithm~S3 of the supplement).

\textbf{A front stage in linear time.} The input streams through a front stage in chunks, so memory holds only what survives. Flows are scored in batches by the detector of Section~\ref{sec:models}, and log records are kept when their level is severe or their message carries an attack indicator (Section~\ref{sec:loganalysis}). The surviving records, called suspects, are merged into incidents with the predicates $P_1$ to $P_3$ of Section~\ref{sec:uicr} in a disjoint set structure. Keys and fingerprints are hashed, and an address in time bucket $\lfloor t/\Delta\rfloor$ is joined to the same address in the previous bucket, so the merge is linear in the number of suspects instead of comparing every record with every open incident as Algorithm~S1 does. Suspects that share an attack signature then form one work unit, whatever their incident. The signature is the detector's attack family for flows, or the message pattern for logs, with the target service and network or host. Each unit receives the triage score of Eq.~(S4) on its evidence, and units are ranked by $\mathrm{TS}(w)\log_{10}(1+|w|)$, where $|w|$ counts its suspects.

\textbf{Checkpoints answered by a policy.} Each work unit becomes one workflow, and a subsystem runs its $N_u$ highest-ranked units on $N_w$ parallel workers. A policy principal approves every checkpoint. It is a system identity naming the subsystem and the analyst who launched it, so every ledger decision stays attributable to a person.

\textbf{Rule buckets instead of deployment.} Approving every checkpoint is safe only because the deployment agent of a subsystem never writes to the rule store. It places every rule, with its engine verdict and the number of suspects it covers, into a bucket kept for each subsystem and attack family. Analysts or the SOC team review the buckets when convenient, by item or in bulk. An approval deploys the rule through the deployment agent and is recorded with the analyst's identity in the ledger of the workflow that wrote the rule, and a rule its engine rejected needs an explicit override. A subsystem therefore automates analysis, not change.

\subsection{Carbon-Aware Operation}\label{sec:carbon}

Every model invocation in LanG passes through one tracker, which takes energy from the most precise source available (Fig.~S2 of the supplement): NVML power at 10\,Hz on NVIDIA GPUs, RAPL counters on Intel CPUs, a utilization model over the nominal CPU package power, or, when nothing can be read, an analytical estimate from two floating-point operations per parameter per token \cite{kaplan2020scaling}, calibrated against hardware readings of the same host class (Eq.~(S5) of the supplement). A GPU that idles at tens of watts dominates a call of a few hundred milliseconds, so LanG stores both the total device energy $E_{\mathrm{tot}}$, following CodeCarbon \cite{codecarbon}, and the marginal energy $E_{\mathrm{marg}}$ of a call of duration $T$,
\begin{equation}\label{eq:energy}
E_{\mathrm{tot}}=\int_0^T P(t)\,dt,\qquad E_{\mathrm{marg}}=E_{\mathrm{tot}}-P_{\mathrm{idle}}T,
\end{equation}
where the power $P$ is read from NVML or RAPL or, under the utilization model, is $P=P_{\mathrm{nom}}(0.3+0.7u)$ for CPU utilization $u$ and nominal package power $P_{\mathrm{nom}}$ (65\,W by default), and $P_{\mathrm{idle}}$ is the median power over five quiet seconds at zero GPU utilization, refreshed in the background and stored with each record, since power sampled just before a call still carries the tail of the previous one. The total is what keeping the model ready costs, the marginal one what a new call adds. Emissions follow as
\begin{equation}\label{eq:co2}
\mathrm{CO_2e}=E\cdot\mathrm{PUE}\cdot\kappa_g,
\end{equation}
where $E$ is $E_{\mathrm{tot}}$ or $E_{\mathrm{marg}}$, PUE is the power usage effectiveness of the data center, the ratio of its total facility energy to the energy of its computing equipment, and $\kappa_g$ is the operational carbon intensity of region $g$, taken from a static table of 56 regions built from Ember annual averages \cite{ember2024review}, which keeps results reproducible and lets the platform run air-gapped. Each record carries a tenant, a model, an operation label and a success flag, a tenant administrator sets a monthly budget with an alert level, 80\% by default, and a telemetry fault removes a record but never blocks an inference. Algorithm~S4 of the supplement summarizes the tracker.

\section{Evaluation}\label{sec:eval}

Sections~\ref{sec:evalcorr} and \ref{sec:evalrules} compare it with the previous LanG, Section~\ref{sec:evalscale} runs it at scale, Sections~\ref{sec:evalnet} to \ref{sec:evalft} compare it with Claude Opus 5.5 (first question), Section~\ref{sec:evalsec} attacks its safeguards (second question), and Section~\ref{sec:effort} measures energy (third question).

\subsection{Setup}

All experiments ran on one workstation with an NVIDIA GeForce RTX 4080 SUPER GPU, Python 3 and Ollama for local inference. For correlation and analysis we used Llama~3.2 with 3B parameters and Qwen~2.5 with 7B parameters, both with 4-bit quantization. Carbon figures use the world average intensity of 475\,gCO$_2$e/kWh and a PUE of 1.0 unless stated otherwise.

\textbf{Corpora.} We use six log corpora, each generated to exercise one ATT\&CK technique: remote system discovery (T1018), brute force (T1110), account manipulation (T1098), data from the local system (T1005), web protocols (T1071) and exfiltration (T1041). Each corpus is a JSON lines file of gateway and application logs, labeled per record with the technique of its generator, whereas our earlier benchmarks took every error or warning line as an attack.

\textbf{Benchmarks.} Benchmark A samples 2{,}000 events from each corpus and defines ground-truth incidents as attack events that share a source address within a 10-minute window. Benchmark B tests correlation across files with $M$ overlapping campaigns, each a six-stage kill chain whose stages come from the six corpora, so every campaign spans six log files. Network stages carry the attacker, victim and command and control addresses. Endpoint stages carry only host and account, except for a fraction $\phi$ of events that also record the host address, as logon events do. Campaigns start 12 minutes apart and stages 9 minutes apart, so they interleave, and benign records reuse the same hosts, accounts and subnets. A fraction $\eta$ of campaigns share one egress address, as behind NAT. The grid crosses $M$$\in\{4,8\}$, 150 or 600 benign events, $\phi\in\{0,0.3,0.6\}$ and $\eta$$\in\{0,0.5,1\}$, 30 seeds each.

\textbf{Real alerts.} GUIDE \cite{freitas2024guide} holds 13 million pieces of evidence from Microsoft Defender XDR across one million incidents, grouped by its production correlation and graded by customer analysts. We aggregate its training file into 1{,}265{,}644 alerts of 5{,}769 organizations, treat placeholder codes of entity columns as missing, and draw 50 windows of one day (207 to 1{,}348 alerts, at least 20 organizations each) and 20 windows of seven days, whose ground truth, the pair of organization and incident, is Defender's own grouping by shared entities. The AIT alert data set \cite{landauer2024alert} holds Suricata, Wazuh and AMiner alerts of eight simulated organizations, with the time window of every attack step. An alert takes its window's step only if its rule fires at least ten times its background rate there, split by organization (five to train, one to validate, two to test).

\textbf{Statistics.} Proportions carry Wilson intervals (Clopper and Pearson at 0 or $n$), other metrics bias-corrected and accelerated bootstrap intervals that resample whole clusters (seeds, windows, organizations, networks, or the alerts of one rule in one scenario) or, for the three test networks of the family namer, single flows, paired proportions the exact McNemar test \cite{mcnemar1947}, and families of comparisons the Holm correction. All intervals are at 95\%. The code and results will be made public at \url{https://github.com/RoronoaZ/LanG} (Section~S13 of the supplement).

\textbf{Detection and rule generation setup.} The detection experiments use the NetFlow~v3 releases \cite{luay2026time} of UNSW-NB15, ToN-IoT, BoT-IoT and CSE-CIC-IDS2018 \cite{moustafa2015unsw,moustafa2021toniot,koroniotis2019botiot,sharafaldin2018cicids}, CICFlowMeter exports of ToN-IoT, BoT-IoT and UNSW-NB15 (the last named CIC-UNBW24), and BCCC-Cloud-DDoS24 exported by NTLFlowLyzer: eight datasets from five networks and three exporters. Detection runs on the CPU (Intel i7-12700KF). Its RAPL counters need root privileges, which we do not grant, so its energy comes from the utilization model of Eq.~\eqref{eq:energy}, with an upper bound at the CPU's full 190\,W turbo power. Rule generation and the local language models run on the GPU, metered by NVML and timed in a separate pass with no other load. The frontier baseline is Claude Opus 5.5, queried once per item without tools, with 0.24 to 0.34\,Wh per request from published figures \cite{elsworth2025google,altman2025gentle,you2025epoch}. Rule engines are Snort~2.9.15, Snort~3.10, Suricata~6.0.4 and~7.0.17, YARA~4.1 and pySigma~2.0.

\textbf{Baselines and metrics.} We compare with union by shared address (IP only), union by shared address, host or account (entity union, which mimics pivoting in a SIEM), a union by shared entities that drops entities seen in more events than a threshold chosen on held-out windows (40 on Benchmark B, 320 on GUIDE), inspired by GraphWeaver \cite{freitas2024graphweaver} (hub-pruned union), and a time window partition with a 5-minute gap. Unanchored hooks is the graph of eight hooks and Louvain stage exactly as released in our previous work \cite{abdennebi2026lang}, and LanG anchored differs from it only by Eq.~\eqref{eq:anchor}. We report B$^3$ precision, recall and F1 over attack events \cite{bagga1998bcubed,amigo2009comparison} and noise isolation, the fraction of predicted attack clusters that contain no benign event.

\subsection{Correlation Quality}\label{sec:evalcorr}

B$^3$ scores each attack event: precision is the share of its predicted group from its own campaign, recall the share of its campaign in its group, both averaged over events, and noise isolation is the share of predicted groups free of benign events. On Benchmark A, whose incidents are defined by address and time, union by address and LanG reach a B$^3$ F1 of 99.90\% to 99.97\% on all six corpora with full noise isolation, while the time window baseline merges everything (0.10\%), so it cannot rank methods linking by address.

\begin{table}[t]
\centering
\caption{Correlation quality (B$^3$, \%, attack events). Benchmark B: mean over 12 cells of $M$, benign volume and $\phi$ (30 seeds each), without ($\eta=0$) and with ($\eta=1$) a shared egress address. GUIDE: 50 one-day and 20 seven-day windows. NI: noise isolation.}
\label{tab:multifile}
\scriptsize
\setlength{\tabcolsep}{1.7pt}
\begin{tabular*}{\columnwidth}{@{\extracolsep{\fill}}l cccc c ccc c@{}}
\toprule
& \multicolumn{5}{c}{Benchmark B} & \multicolumn{4}{c}{GUIDE} \\
\cmidrule(lr){2-6}\cmidrule(lr){7-10}
& \multicolumn{4}{c}{$\eta=0$} & $\eta=1$ & \multicolumn{3}{c}{one day} & 7 days \\
\cmidrule(lr){2-5}\cmidrule(lr){6-6}\cmidrule(lr){7-9}\cmidrule(lr){10-10}
Method & P & R & F1 & NI & F1 & P & R & F1 & F1 \\
\midrule
Time window & 18.80 & 99.88 & 31.17 & 0.00 & 31.13 & 42.68 & 48.57 & 28.82 & 59.78 \\
IP only & 99.67 & 44.83 & 60.12 & 99.65 & \textbf{42.12} & 84.38 & 59.87 & \textbf{60.41} & \textbf{71.58} \\
Entity union & 39.06 & 83.33 & 45.42 & 29.32 & 30.77 & 41.68 & 85.47 & 45.70 & 41.81 \\
Hub-pruned union & 42.10 & 78.77 & 48.21 & 47.48 & 40.82 & 46.08 & 82.53 & 47.43 & 44.10 \\
Unanchored hooks & 18.75 & 59.10 & 27.91 & 13.98 & 27.45 & 28.52 & 76.15 & 25.16 & 13.21 \\
\textbf{LanG anchored} & 51.81 & 51.32 & 49.30 & 73.03 & 24.49 & 45.09 & 68.72 & 37.04 & 42.18 \\
\quad without kill chain & 63.43 & 62.45 & \textbf{61.29} & 68.19 & 26.46 & 45.47 & 67.50 & 37.37 & 43.07 \\
\quad without host, user & 48.15 & 46.28 & 44.99 & 73.66 & 23.66 & 55.86 & 64.50 & 40.35 & 55.89 \\
\bottomrule
\end{tabular*}
\end{table}

Table~\ref{tab:multifile} reports the harder Benchmark B and the real GUIDE alerts, with three findings. First, anchoring fixes the phase collapse of our previous graph. Without it, precision on Benchmark B stays at 18.75\%, the share of one campaign among the attack events, so every group mixes all campaigns, and on GUIDE it reaches only 28.52\% in one day (F1 of 13.21\% over seven days). Requiring a shared entity on every edge (Eq.~\eqref{eq:anchor}) raises precision to 51.81\% and 45.09\%, F1 from 27.91\% to 49.30\% and from 25.16\% to 37.04\% (paired difference 11.88 points, Holm $p=0.005$ over the GUIDE windows), and noise isolation from 13.98\% to 73.03\%. Second, anchoring repairs the graph but does not beat simple grouping: union by address is best on GUIDE (F1 of 60.41\% in one day, 71.58\% over seven days), whose truth is Defender's own grouping by shared entities, and ties the best anchored graph on Benchmark B (60.12\% against 61.29\% without the kill chain hook, which hurts when all campaigns share their phases). Third, an address shared by several campaigns, as behind NAT, misleads every method that links through addresses: with $\eta=1$, union by address falls to 42.12\% and the anchored graph to 24.49\%, since that address counts as a shared entity. Ignoring entities that occur in too many events (hub pruning) helps: the pruned union degrades least among the unions using hosts and accounts, and inside the anchored graph, at the same threshold, pruning raises F1 to 41.18\% (significant in 10 of 12 cells), or 52.08\% without the kill chain hook, above union by address, at no cost otherwise. On GUIDE it adds 2.35 points (39.39\%), not significantly. Host and network evidence that share nothing still need an asset inventory mapping hosts to addresses, or causal evidence such as a victim logon. The graph of 100 events takes 0.13\,s to build and partition, growing quadratically (Section~S5 of the supplement).

\subsection{Autonomous Subsystems at Scale}\label{sec:evalscale}

On our single workstation, we launched two autonomous subsystems at once on 3.33 million records, each executing the workflow of Fig.~\ref{fig:agentic}. The first analyzed the complete NetFlow~v3 export of UNSW-NB15, 2{,}365{,}424 flows, with a detector retrained without UNSW-NB15, so that the network was new to it, and a baseline of 5{,}000 of its normal flows. The second analyzed the 965{,}067 lines of the six log corpora. Each ran its 25 highest-ranked units ($N_u=25$) on four workers ($N_w=4$), with Llama~3.2 3B and the served rule generator, eight workflows at a time on one GPU. Labels were used only for scoring.

Table~S1 of the supplement details the runs. The front stage processed about 46{,}000 flows and 53{,}000 log lines per second. Both subsystems finished in 382.2\,s, all 50 workflows completed, and their policies answered 100 checkpoints. The 25 units of each subsystem held 93.17\% of the suspected attack flows, with all nine attack classes of UNSW-NB15, and 97.20\% of the suspected attack lines, with both ATT\&CK techniques present. The subsystems left 207 rules in six buckets, of which 201 loaded in their engines, and deployed none. One workflow per suspect means 200{,}114 workflows, or about two weeks at the measured rate, against 6.4 minutes. The devices used 25.99\,Wh, or 0.52\,Wh per workflow, while the emission records of the individual calls sum to 5.45 times more, because concurrent calls on one GPU overlap and each record counts the whole device power during its call. Under concurrency, energy must therefore be measured over a window for the whole device. The filter kept only 68.37\% of the attack log lines, because many discovery and account manipulation events are logged at the INFO level without an attack keyword, so rules over log content must complement it.

\subsection{Detection Across Networks}\label{sec:evalnet}

\begin{figure}[t]
\centering
\begin{tikzpicture}[font=\scriptsize]
\pgfplotsset{netax/.style={scale only axis, y dir=reverse, xmajorgrids, grid style={black!12}, axis line style={black!50},
  tick label style={font=\tiny}, label style={font=\scriptsize}, ytick style={draw=none}, title style={font=\scriptsize, yshift=-1.2ex}},
  ci/.style={only marks, error bars/.cd, x dir=both, x explicit, error bar style={line width=0.45pt, black!70}, error mark options={rotate=90, mark size=1.1pt, black!70}}}
\begin{axis}[netax, name=a, width=0.52\columnwidth, height=1.55cm, xmin=0, xmax=1.02, ymin=0.5, ymax=4.5,
  ytick={1,...,4}, yticklabels={BoT-IoT,CSE-CIC-IDS2018,ToN-IoT,UNSW-NB15}, xtick={0,0.25,0.5,0.75,1}, xticklabels={0,0.25,0.5,0.75,1},
  title={(a) Gradient boosted trees and site baseline: AUC on each NetFlow dataset},
  legend style={font=\tiny, at={(1.03,0.5)}, anchor=west, draw=none, cells={anchor=west}, row sep=-1pt}]
\draw[black!35, densely dashed] (axis cs:0.5,0.5) -- (axis cs:0.5,4.5);
\addplot[only marks, mark=o, mark size=1.7pt, black, error bars/.cd, x dir=both, x explicit, error bar style={line width=0.45pt, black!45}, error mark options={rotate=90, mark size=1.1pt, black!45}] coordinates {(0.549,1) += (0.213,0) -= (0.406,0) (0.353,2) += (0.371,0) -= (0.122,0) (0.820,3) += (0.000,0) -= (0.249,0) (0.396,4) += (0.038,0) -= (0.222,0)};
\addlegendentry{trees, other networks}
\addplot[only marks, mark=square*, mark size=1.5pt, black!55] coordinates {(0.996,1) (0.964,2) (0.947,3) (1.000,4)};
\addlegendentry{trees, same network}
\addplot[only marks, mark=triangle*, mark size=2.0pt, black] coordinates {(0.871,1) (0.885,2) (0.854,3) (0.929,4)};
\addlegendentry{site baseline, no labels}
\end{axis}
\begin{axis}[netax, name=b, at={(a.south west)}, anchor=north west, yshift=-0.85cm, width=0.30\columnwidth, height=3.0cm,
  xmin=0.5, xmax=1, ymin=0.4, ymax=13.6, ytick={1,...,13}, yticklabels={{\textbf{LanG baseline}},{Nearest neighbors, raw},{One class SVM, raw},{One class SVM, ranks},{Autoencoder, raw},{KitNET, raw},{LOF, ranks},{Isolation forest, ranks},{KitNET, ranks},{LOF, raw},{Autoencoder, ranks},{Isolation forest, raw},{Rank tail (ECOD)}},
  xtick={0.5,0.75,1}, xticklabels={0.5,0.75,1}, title={(b) Mean AUC}]
\addplot[mark=o, mark size=1.5pt, black, ci] coordinates {(0.93,2) += (0.03,0) -= (0.04,0) (0.69,3) += (0.11,0) -= (0.13,0) (0.74,4) += (0.11,0) -= (0.10,0) (0.88,5) += (0.04,0) -= (0.09,0) (0.82,6) += (0.07,0) -= (0.11,0) (0.86,7) += (0.05,0) -= (0.06,0) (0.89,8) += (0.04,0) -= (0.04,0) (0.75,9) += (0.08,0) -= (0.07,0) (0.83,10) += (0.06,0) -= (0.05,0) (0.84,11) += (0.04,0) -= (0.10,0) (0.88,12) += (0.04,0) -= (0.05,0) (0.78,13) += (0.09,0) -= (0.12,0)};
\addplot[mark=*, mark size=1.8pt, black, ci] coordinates {(0.93,1) += (0.03,0) -= (0.05,0)};
\end{axis}
\begin{axis}[netax, name=c, at={(b.north east)}, anchor=north west, xshift=0.25cm, width=0.30\columnwidth, height=3.0cm,
  xmin=0, xmax=75, ymin=0.4, ymax=13.6, ytick={1,...,13}, yticklabels={}, xtick={0,25,50,75},
  title={(c) Attacks detected (\%)}]
\addplot[mark=o, mark size=1.5pt, black, ci] coordinates {(38.38,2) += (25.33,0) -= (20.27,0) (35.73,3) += (24.85,0) -= (18.59,0) (32.86,4) += (24.98,0) -= (17.59,0) (32.12,5) += (25.35,0) -= (17.18,0) (20.77,6) += (25.19,0) -= (10.11,0) (15.93,7) += (18.32,0) -= (7.38,0) (13.67,8) += (9.68,0) -= (7.36,0) (13.42,9) += (11.25,0) -= (6.21,0) (13.11,10) += (17.77,0) -= (7.09,0) (13.07,11) += (8.78,0) -= (6.18,0) (9.90,12) += (6.82,0) -= (5.51,0) (8.47,13) += (5.33,0) -= (4.49,0)};
\addplot[mark=*, mark size=1.8pt, black, ci] coordinates {(42.66,1) += (24.07,0) -= (21.48,0)};
\end{axis}
\end{tikzpicture}
\caption{Detection on unseen networks. (a) Trees trained on other NetFlow datasets (bars: eight feature variants) are near or below chance (dashed) except on ToN-IoT, unlike trees trained on the same dataset and the label-free site baseline. (b, c) Label-free detectors at 1\% false alarms, means over 40 runs with 95\% intervals.}
\label{fig:net}
\end{figure}
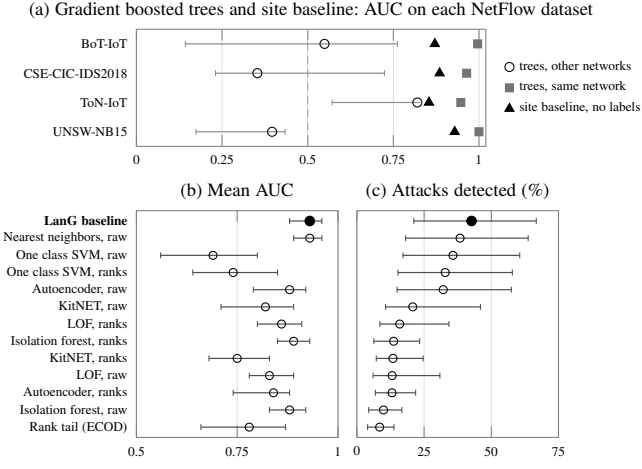

\textbf{Supervised models do not transfer.} We trained gradient boosted trees on three of the four NetFlow~v3 datasets and tested them on the fourth, holding out each dataset in turn (Fig.~\ref{fig:net}a). With all 47 NetFlow features, the area under the ROC curve (AUC) is only 0.35 to 0.82 and the family macro F1 3.88\% to 17.26\%, and removing the TTL and port features, keeping only the shared fields or scaling ranks per dataset does not help. On separate time blocks of the same dataset, the same model reaches a family macro F1 of 59.29\% to 95.68\%. What transfers is a model of the target network's own benign traffic, with an AUC of 0.80 to 0.96 in every variant and no attack label, motivating Section~\ref{sec:models}.

\begin{table}[t]
\centering
\caption{Detector on each dataset as an unseen network. Left: with 5{,}000 normal flows only, AUC, attacks detected and benign flows flagged (\%) at 1.00\% target false alarms. Right: family macro F1 with $n$ analyst labels. NF, CIC, NTL: NetFlow~v3, CICFlowMeter, NTLFlowLyzer.}
\label{tab:sites}
\scriptsize
\setlength{\tabcolsep}{2.4pt}
\begin{tabular*}{\columnwidth}{@{\extracolsep{\fill}}l ccc ccc@{}}
\toprule
& \multicolumn{3}{c}{No labels from the network} & \multicolumn{3}{c}{Family macro F1 (\%)} \\
\cmidrule(lr){2-4}\cmidrule(lr){5-7}
Dataset (export) & AUC & Detected\,\% & False al.\,\% & $n{=}0$ & $200$ & $1000$ \\
\midrule
UNSW-NB15 (NF) & 0.997 & 92.48 & 0.88 & 6.94 & 51.01 & 50.89 \\
UNSW-NB15 (CIC) & 0.914 & 35.61 & 1.13 & 9.24 & 47.29 & 51.93 \\
ToN-IoT (NF) & 0.881 & 8.54 & 1.10 & 4.17 & 79.82 & 85.71 \\
ToN-IoT (CIC) & 0.896 & 5.13 & 1.22 & 3.83 & 46.29 & 53.75 \\
BoT-IoT (NF) & 0.968 & 87.71 & 1.31 & 8.51 & 88.48 & 90.09 \\
BoT-IoT (CIC) & 0.973 & 63.53 & 1.00 & 11.59 & 82.01 & 87.19 \\
CSE-CIC-IDS2018 (NF) & 0.944 & 34.87 & 0.78 & 4.47 & 62.06 & 69.17 \\
Cloud-DDoS24 (NTL) & 0.770 & 0.27 & 0.66 & 27.04 & 86.13 & 87.85 \\
\bottomrule
\end{tabular*}
\end{table}

\textbf{The false alarm rate stays predictable.} Table~\ref{tab:sites} treats each dataset as a new customer network. In the left half, the detector flags a flow when it lies further from the 5{,}000 normal flows than 99.00\% of normal flows do. On every dataset, 0.66\% to 1.31\% of the benign flows are flagged, close to the 1.00\% target, so an operator can predict the alarm volume on an unseen network. How many attacks are caught at that rate depends on how different they look from normal traffic: 87.71\% and 92.48\% of the attack flows on the NetFlow exports of BoT-IoT and UNSW-NB15, but only 0.27\% of the cloud DDoS flows, which imitate normal requests.

\textbf{The baseline beats other label-free detectors.} Fig.~\ref{fig:net}b and c compare the detector with seven label-free detectors fitted on the same 5{,}000 normal flows and thresholded at the same 1\% false alarm target, each on percentile ranks and on raw values: isolation forest \cite{liu2008isolation}, local outlier factor (LOF), one-class SVM, an autoencoder, the KitNET ensemble of Kitsune \cite{mirsky2018kitsune} on flow features, a rank tail score in the manner of ECOD \cite{li2023ecod} and nearest neighbors on raw values. Over 40 runs (eight data sets, five baseline draws), the detector detects the most attacks, 42.66\% [21.18, 66.73] at a realized false alarm rate of 0.93\%, with a mean AUC of 0.926 [0.872, 0.959], and is the best in 14 of the 40 runs. Nearest neighbors on raw values reach a similar AUC (0.929) but detect fewer attacks (38.38\%), and a high AUC does not ensure detection at a low alarm rate: isolation forest on ranks reaches 0.894 but detects 13.67\%. The autoencoder on raw values detects 32.12\% at 17 times the CPU energy. The detector scores 864{,}341 flows per second, for $2.7\times10^{-3}$\,Wh of CPU energy per run. Averaging its score with isolation forest and the rank tail lowered detection on 7 of 8 data sets (24.38\% against 42.66\%), so the spread comes from the networks, not the scorer.

\textbf{Labels from analysts name the attacks.} Without labels from the network, the family classifier (right half of Table~\ref{tab:sites}) reaches a macro F1 of only 3.83\% to 27.04\%. With 200 analyst labels, it reaches 46.29\% to 88.48\%, and 1{,}000 labels add little more. UNSW-NB15 stays near 50.00\% because several of its labels describe overlapping traffic, and a model trained and tested on UNSW-NB15 alone reaches only 59.29\%.

\textbf{The baseline must be refreshed.} A baseline taken from the first days only flags 47.70\% and 83.80\% of the benign flows of the two ToN-IoT exports, because their normal traffic drifts.

\subsection{Detection Versus Language Models}\label{sec:evalllm}

All systems judged the same 200 flows, 25 benign and 25 attack flows from each NetFlow dataset, removed from every LanG training set. The local language models received each flow's features with units and flag definitions, the family list and a request for a JSON verdict, a long prompt that costs them only power, whereas Claude Opus 5.5 received a minimal prompt, to obtain its best answer at the lowest cost per paid request. In the \emph{unseen} setting, LanG's classifier was trained only on the other networks, and the network contributed normal flows but no labels. The \emph{adapted} setting adds 1{,}000 labeled flows, and the \emph{same-dataset} setting trains on the network's own labels.

\begin{table}[t]
\centering
\caption{Detection and family naming on 200 flows of four networks (\%) and energy per flow (LanG batched and per call, frontier published, inference only, training in Table~\ref{tab:energy}). The frontier prompt differed from the local one. Section~\ref{sec:evalft} compares naming under one protocol.}
\label{tab:llmdet}
\scriptsize
\setlength{\tabcolsep}{2pt}
\begin{tabular*}{\columnwidth}{@{\extracolsep{\fill}}l ccccc c@{}}
\toprule
System & P & R & F1 & FPR & Family & Wh per flow \\
\midrule
LanG, unseen network & \textbf{100.00} & 58.00 & 73.42 & \textbf{0.00} & 4.50 & \multirow{3}{*}{\shortstack{$4.9{\times}10^{-7}$ and\\$9.3{\times}10^{-5}$}} \\
LanG, adapted & 91.95 & 80.00 & 85.56 & 7.00 & 62.57 & \\
LanG, same dataset & 98.84 & \textbf{85.00} & \textbf{91.40} & 1.00 & \textbf{71.32} & \\
\midrule
Claude Opus 5.5 & 55.45 & 61.00 & 58.10 & 49.00 & 43.26 & 0.24 to 0.34 \\
Qwen 2.5 7B & 0.00 & 0.00 & 0.00 & 1.00 & 0.00 & 0.026 \\
Llama 3.2 3B & 0.00 & 0.00 & 0.00 & 0.00 & 2.28 & 0.015 \\
\bottomrule
\end{tabular*}
\end{table}

Table~\ref{tab:llmdet} shows that on an unseen network LanG raises no false alarm (0 of 100 benign flows, a 95\% upper bound of 3.62\%) and finds 58.00\% [48.21, 67.20] of the attacks. The frontier model finds a similar share (61.00\%) but flags 49.00\% [39.42, 58.65] of the benign flows, an alert volume no SOC could triage, and LanG is correct on 68 flows where it errs against 22 the other way (McNemar $p<0.001$). Both local models call nearly every flow benign. The frontier model names the family better than LanG on an unseen network (43.26\% against 4.50\%), but in the pipeline the classification agent sees only the flows the detector flags. With the frontier model naming the families of LanG's alerts, detection stays that of LanG, family accuracy on the detected attacks rises from 17.24\% to 53.45\%, and the language model runs only on alerts. With 1{,}000 labels, LanG alone names families better (62.57\%). However, adaptation raised the false positive rate on CSE-CIC-IDS2018 to 24.00\%, so an adapted model must be checked on held-out normal flows before deployment.

On the CPU, the detector processes 14{,}179 flows per second in batches and takes 6.8\,ms per flow one at a time (14.2\,ms at the 99th percentile), so even at 190\,W a flow costs at least 650 times less energy than a frontier request one at a time and 60{,}000 times less in batches (Table~\ref{tab:energy}). Section~S7 of the supplement reports the retrained text and DNS over HTTPS classifiers.

\subsection{Rule Generation}\label{sec:evalrules}

The benchmark holds 240 test incidents, 40 per target (Snort~2.9, Snort~3, Suricata~6, Suricata~7, YARA and Sigma), whose rule templates never occur in training. Finetuned models receive their training prompt, and the others a generic prompt with the same facts. A rule is \emph{accepted} when the engine of the requested version loads it without repair, and an answer with no rule counts as rejected. For \emph{functional detection}, each incident's indicators are synthesized into traffic (HTTP, DNS, TLS or raw TCP and UDP) and replayed through the engine, and only the 84 incidents whose reference rule fires on this traffic are scored. YARA rules are scanned against a sample built from the indicators, and Sigma rules are evaluated on a synthetic event. Benign alerts are counted on 400 benign flows, 60 system binaries and other incidents.

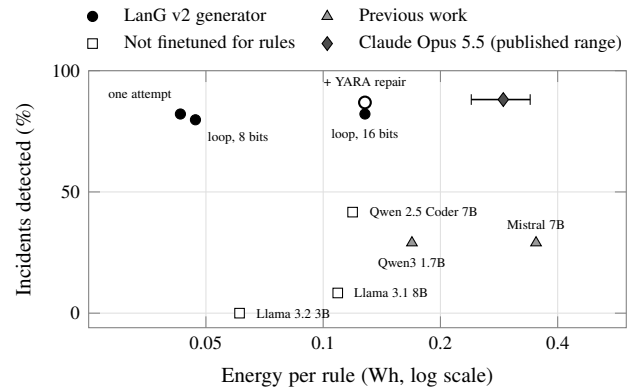
\begin{figure}[t]
\centering
\begin{tikzpicture}
\begin{axis}[
  scale only axis, width=0.80\columnwidth, height=3.4cm,
  xmode=log, xmin=0.025, xmax=0.6, ymin=-6, ymax=100,
  xtick={0.05,0.1,0.2,0.4}, xticklabels={0.05,0.1,0.2,0.4},
  xlabel={Energy per rule (Wh, log scale)}, ylabel={Incidents detected (\%)},
  tick label style={font=\scriptsize}, label style={font=\footnotesize},
  grid=major, grid style={black!12}, axis line style={black!50},
  legend style={font=\scriptsize, at={(0.5,1.03)}, anchor=south, legend columns=2, draw=none, column sep=8pt, cells={anchor=west}},
]
\addplot[only marks, mark=*, mark size=1.9pt, black] coordinates {(0.128,82.14) (0.047,79.76) (0.043,82.14)};
\addlegendentry{LanG v2 generator}
\addplot[only marks, mark=triangle*, mark size=2.3pt, black, mark options={fill=black!40}] coordinates {(0.169,29.03) (0.352,29.03)};
\addlegendentry{Previous work}
\addplot[only marks, mark=square*, mark size=1.8pt, black, mark options={fill=white}] coordinates {(0.119,41.67) (0.109,8.33) (0.061,0.00)};
\addlegendentry{Not finetuned for rules}
\addplot[only marks, mark=diamond*, mark size=2.6pt, black, mark options={fill=black!70}, error bars/.cd, x dir=both, x explicit, error bar style={line width=0.6pt, black}, error mark options={rotate=90, mark size=2pt, black}] coordinates {(0.29,88.10) += (0.05,0) -= (0.05,0)};
\addlegendentry{Claude Opus 5.5 (published range)}
\addplot[only marks, mark=o, mark size=2.2pt, black, thick] coordinates {(0.128,86.90)};
\node[font=\tiny, anchor=south] at (axis cs:0.128,88.6) {+ YARA repair};
\node[font=\tiny, anchor=north] at (axis cs:0.128,79.6) {loop, 16 bits};
\node[font=\tiny, anchor=north west] at (axis cs:0.048,78.6) {loop, 8 bits};
\node[font=\tiny, anchor=south east] at (axis cs:0.043,84.0) {one attempt};
\node[font=\tiny, anchor=north] at (axis cs:0.169,26.5) {Qwen3 1.7B};
\node[font=\tiny, anchor=south] at (axis cs:0.352,31.0) {Mistral 7B};
\node[font=\tiny, anchor=west] at (axis cs:0.125,41.67) {Qwen 2.5 Coder 7B};
\node[font=\tiny, anchor=west] at (axis cs:0.114,8.33) {Llama 3.1 8B};
\node[font=\tiny, anchor=west] at (axis cs:0.064,0.0) {Llama 3.2 3B};
\end{axis}
\end{tikzpicture}
\caption{Rule generation on 240 held-out incidents: replayable incidents detected (84, or 62 without Sigma for the previous work) against energy per rule. The open circle is the loop with the YARA repair. Table~S2 of the supplement gives acceptance, indicator recall and latency.}
\label{fig:rules}
\end{figure}

Fig.~\ref{fig:rules} reports the results (full values in Table~S2 of the supplement). With LanG's validator loop, 98.33\% of the new generator's rules load in their engine and 82.14\% detect their incident, with no benign alert, against 57.50\% to 60.00\% and 29.03\% for the generators used in our previous work \cite{abdennebi2026lang}. Paired on the same incidents, it is not significantly different from the frontier model on acceptance ($p=0.38$, 4 incidents favoring the frontier model against 1) and detection ($p=0.36$, 12 against 7), although the frontier model is ahead numerically. The paired differences are $-1.25$ points [$-3.82$, 0.92] on acceptance and $-5.95$ points [$-16.54$, 4.63] on detection, so with 84 replayable incidents a gap of up to 16 points cannot be ruled out. A single attempt is significantly worse on acceptance ($p=0.006$), so the workflow always runs the loop, at 1.07 to 1.10 attempts per rule. Network rules are on par with the frontier model, and the gap lies in YARA (66.67\% against 80.00\% detection), where the generator adds unsupported constraints such as wide strings, PE header checks and size limits, and in Sigma (90.91\% against 100.00\%). A deterministic repair, now part of the validator loop, adds \texttt{ascii} to strings marked \texttt{wide} and drops PE header and file size tests that the incident does not state. Applied to the saved rules, it raises detection to 86.90\% (73 of 84, 14 of 15 YARA incidents) for the loop and to 83.33\% for the served generator, without benign alerts, against 88.10\% for the frontier model (McNemar $p=1.00$). On 360 further YARA incidents, never used to design it, it raises detection from 57.41\% [47.99, 66.32] to 80.56\% [72.10, 86.92] of the 108 replayable ones (McNemar $p<0.001$), with the same 4 benign alerts and two fewer rules that load. The best local model not finetuned for rules accepts 54.17\% and detects 41.67\%.

Serving the same weights through llama.cpp at 8 bits instead of Python at 16 bits leaves quality within sampling noise, keeps the loop not significantly different from the frontier model on acceptance and detection ($p=0.13$ and $0.19$), and cuts the energy per attempt from 0.120 to 0.043\,Wh and the median latency from 2.86 to 0.78\,s. The generator then uses the least energy of all systems (Fig.~\ref{fig:rules}), 2.5 to 2.8 times less than the Ollama 7B and 8B models and 5.6 to 7.9 times less than a frontier request, and 0.05 to 0.06\,Wh per rule that loads and detects (Table~\ref{tab:energy}), against 0.28\,Wh for Qwen~2.5 Coder 7B. Training cost at most 1.1\,kWh, repaid by frontier savings after 3{,}800 to 5{,}700 rules.

\subsection{Finetuned Agents Versus the Frontier Model}\label{sec:evalft}

Our goal is to measure how far small models that we train and run locally can go, with frontier models as a reference rather than competitors. Our subscription budget limits that reference to one hosted model, Claude Opus 5.5, queried once per item, since querying several frontier models is beyond our current budget and many cap query volumes, and to the open-weight gpt-oss-20b \cite{openai2025gptoss}, run on our GPU. We registered data, items, endpoints and decision rules before any test inference. Every system receives the same instruction and item, Claude Opus 5.5 answers each item in a fresh context without tools, and untuned local models run with thinking off, except gpt-oss-20b, which reasons at low effort. A system \emph{outperforms} another if the Holm-adjusted $p$ is below 0.05 and the 95\% interval of the paired macro F1 difference excludes zero, and two systems are \emph{equivalent} if that interval lies within $\pm5$ points. Within our budget, the frontier model answered a stratified subset of each test set (397 alerts, 250 incidents, 217 flows) rather than all 18{,}792 items, and the untuned local models answered the same subsets, so every comparison with it is paired. The finetuned and classical systems were also scored on the whole test sets, and keyword rules exist for alert steps only (Tables~S3 and S4 of the supplement give all values and the escalation ladder).

\begin{figure}[t]
\centering
\begin{tikzpicture}
\begin{axis}[
  scale only axis, width=0.66\columnwidth, height=4.3cm,
  xmin=0, xmax=100, ymin=0.45, ymax=8.55, y dir=reverse,
  ytick={1,...,8}, yticklabels={\textbf{LanG finetuned},Claude Opus 5.5,gpt-oss-20b,Classical,Keyword rules,Qwen 2.5 7B,Qwen 3.5 4B,Llama 3.2 3B},
  xtick={0,20,40,60,80,100}, xlabel={Macro F1 (\%) with 95\% interval},
  tick label style={font=\scriptsize}, label style={font=\footnotesize},
  xmajorgrids, grid style={black!12}, ytick style={draw=none}, axis line style={black!50},
  extra y ticks={1.5,2.5,3.5,4.5,5.5,6.5,7.5}, extra y tick labels={}, extra y tick style={grid=major, major grid style={black!8}, tick style={draw=none}},
  legend style={font=\scriptsize, at={(0.5,1.02)}, anchor=south, legend columns=3, draw=none, column sep=6pt},
]
\addplot[only marks, mark=*, mark size=1.7pt, black, error bars/.cd, x dir=both, x explicit, error bar style={line width=0.45pt, black!70}, error mark options={rotate=90, mark size=1.2pt, black!70}] coordinates {(97.31,0.76) += (2.09,0) -= (4.50,0) (95.19,1.76) += (2.64,0) -= (4.46,0) (53.64,2.76) += (8.69,0) -= (7.65,0) (98.32,3.76) += (1.27,0) -= (19.89,0) (56.00,4.76) += (6.09,0) -= (5.93,0) (36.83,5.76) += (14.89,0) -= (8.99,0) (37.98,6.76) += (7.73,0) -= (6.18,0) (18.73,7.76) += (4.62,0) -= (4.24,0)};
\addlegendentry{Alert step}
\addplot[only marks, mark=square*, mark size=1.6pt, black, fill=white, mark options={fill=white}, error bars/.cd, x dir=both, x explicit, error bar style={line width=0.45pt, black!70}, error mark options={rotate=90, mark size=1.2pt, black!70}] coordinates {(42.83,1.00) += (13.42,0) -= (18.18,0) (27.24,2.00) += (7.92,0) -= (6.98,0) (30.07,3.00) += (13.12,0) -= (8.06,0) (43.41,4.00) += (12.70,0) -= (16.36,0) (23.42,6.00) += (13.15,0) -= (7.94,0) (18.49,7.00) += (9.36,0) -= (6.49,0) (19.18,8.00) += (8.98,0) -= (6.76,0)};
\addlegendentry{Incident grade}
\addplot[only marks, mark=triangle*, mark size=2.0pt, black, mark options={fill=black!40}, error bars/.cd, x dir=both, x explicit, error bar style={line width=0.45pt, black!70}, error mark options={rotate=90, mark size=1.2pt, black!70}] coordinates {(12.14,1.24) += (4.11,0) -= (3.14,0) (26.78,2.24) += (6.36,0) -= (5.65,0) (14.78,3.24) += (6.65,0) -= (3.75,0) (12.24,4.24) += (3.42,0) -= (2.64,0) (11.91,6.24) += (3.10,0) -= (2.56,0) (10.64,7.24) += (3.42,0) -= (2.68,0) (7.32,8.24) += (3.77,0) -= (2.46,0)};
\addlegendentry{Flow family}
\end{axis}
\end{tikzpicture}
\caption{Macro F1 of the detection phase agents on the items the frontier model answered (397 alerts, 250 incidents, 217 flows), with 95\% intervals. Classical: TF-IDF with logistic regression (alerts), LightGBM (incidents, flows). Table~S3 of the supplement lists all values.}
\label{fig:ft}
\end{figure}
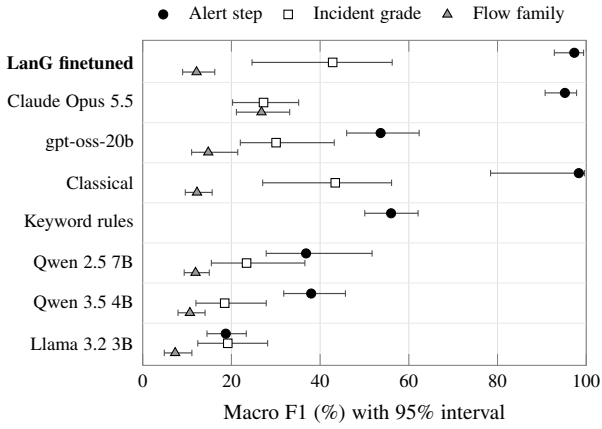

\textbf{Alert classification.} The 4B alert analyst (1.7B stayed below the frontier model on validation and 8B did worse) reaches 97.31\% [92.81, 99.40] on the 397 alerts against 95.19\% [90.73, 97.83] for the frontier model, a difference of 2.11 points [$-0.10$, 4.81] within the equivalence margin, and it is right on 11 alerts where the frontier model errs, against 1 (McNemar $p=0.006$). On all 10{,}297 test alerts it labels 99.15\% correctly (macro F1 90.75\%, lowered by rare privilege escalation steps). TF-IDF with logistic regression is as good (98.32\% and 93.10\%), since an alert's rule and log line nearly determine its step, so our decision rule recommends it. Keyword matching reaches 56.00\%, untuned local models of 3B to 7B parameters at most 37.98\%.

\textbf{Incident grading.} The 1.7B incident grader beat the frontier model on validation and stays at 1.7B. On the 250 incidents it reaches 42.83\% [24.65, 56.25] against 27.24\% [20.26, 35.16], as the frontier model almost never grades an incident a false positive, but the difference (15.59 points [$-0.30$, 29.68]) is not significant. LightGBM is as good (43.41\%, and 42.16\% against 40.95\% on all 3{,}000 incidents) and is the recommended grader, and untuned local models of 3B to 7B parameters reach at most 23.42\%. GUIDE's authors report 0.87 to 0.91 \cite{freitas2024guide} on a split where every test organization appears in training. Ours removes each organization's base rates, as for a new tenant, and 10 labeled incidents per organization restore 69.53\% (77.09\% with 25, against 28.25\%) on the 17 largest test organizations.

\textbf{Naming attack families.} On a new network the frontier model wins. The family namer escalated to 8B (17.52\% at 1.7B and 20.75\% at 4B against 52.61\% on development flows) and reaches 12.14\% [9.00, 16.25] on the 217 flows against 26.78\% [21.13, 33.14], a gap of 14.64 points [8.20, 21.59] (Holm $p<0.001$). On all 5{,}495 flows it beats LightGBM (16.27\% against 12.58\%), within the margin. Site labels close the gap: with 50 labels of the new network, LightGBM names its other flows at 55.07\% (67.08\% with 200), while frontier names make poor labels (23.33\%). LanG therefore uses the frontier model on a new network until 50 alerts are labeled (Table~S10 of the supplement).

\textbf{Open weight reference.} On our GPU, gpt-oss-20b reaches 53.64\%, 30.07\% and 14.78\% on the three subsets. The alert analyst beats it by 43.67 points [35.84, 51.33], the grader and the namer do not differ significantly from it, and the frontier model names flow families 12.00 points [5.69, 18.80] better. It uses 0.032 to 0.054\,Wh per item, two to six times our adapters.

\subsection{Safeguards Under Attack}\label{sec:evalsec}

We attack the three surfaces that LanG adds to a SOC (Section~\ref{sec:model}): telemetry text the agents read, analyst labels retraining the family classifier, and the action ledger. Injections from published corpora enter through the two fields that reach the models, a user agent and a log line. Adaptive and multistep injection, tool abuse, model extraction and denial of service are future work.

\textbf{Indirect prompt injection.} On their own, the input filters catch little. The pattern filter flags 5.32\% of the 263 deepset injections \cite{deepset2023injections} and none of the 150 BIPIA text attacks \cite{yi2025bipia}, and adding Prompt Guard~2 raises these shares to 26.24\% and 0.67\% at 7 to 11\,ms per field, while it flags one of 399 benign prompts (0.25\%) and none of 4{,}000 AIT alerts and log lines (95\% upper bound 0.18\%, Table~S6 of the supplement). BIPIA attacks ask for ordinary tasks that a classifier cannot tell apart from data, so filters cannot be the main defense. In the complete pipeline, we appended 40 BIPIA attacks to a packet user agent or a log line of 40 AIT incidents (30 attacks, 10 benign), approved the classification checkpoint as a rubber-stamp analyst would, and ran to the verdict without filters, with the pattern filter, and with Prompt Guard~2 and spotlighting (120 workflows), with benign deepset prompts as a control. Without spotlighting, the attack reached a model in 62.50\% of the workflows and no filter flagged it. Yet no verdict changed, since the verdict follows a detector that reads no text, no attack incident was graded benign (0 of 90, 95\% upper bound 4.09\%), and every workflow reached the analyst. Severities and log risk levels changed as often as when benign text replaced the attacks or the clean incidents simply ran again (Table~S7 of the supplement), a run-to-run variability of the 3B classification model that came from sampling. With greedy decoding and a fixed seed for every decision-bearing call, a clean rerun changes no severity (0 of 40, against 17 before), and the attacks change severities exactly as often as benign text (10.00\%) and log risk levels no more often than a clean rerun (22.50\% against 25.00\%). An attacker adapting to LanG remains a risk.

\textbf{Poisoned feedback.} We adapted the family classifier with 300 labels (200 attack, 100 benign flows) of each NetFlow data set as a new site, over three baseline draws (12 runs), scored as deployed: the baseline raises an alert and the classifier names it. Clean labels raise detection from 59.48\% to 89.24\% and the family macro F1 from 0.86\% to 61.01\%, at a false alarm rate of 2.57\% instead of 1.08\%. An adversary controlling a fraction $\epsilon$ of the labels who relabels the site's most frequent attack family as benign lowers the share of that family named correctly from 80.97\% to 55.95\% at $\epsilon=5\%$ and 10.97\% at $\epsilon=20\%$, but 64.12\% of its flows still raise an alert, as many as without labels (63.54\%), because the baseline decides what is abnormal. Relabeling benign flows as attacks raises the false alarm rate to at most 2.91\% up to $\epsilon=10\%$ and 7.79\% at $\epsilon=20\%$. The holdout gate rejected half of these adaptations at $\epsilon=20\%$ but no targeted one, whose held-out labels are poisoned too. The label audit sent 63.89\% to 70.83\% of the targeted poisoned labels to a second analyst while flagging 1.17 of 300 clean labels on average. Labels approved by an autonomous policy repeat the current model's errors (macro F1 0.59\%) and are excluded by the principal filter (Table~S8 of the supplement).

\textbf{Ledger tampering.} On copies of the ledger of the 50 workflows of Section~\ref{sec:evalscale} (3{,}052 actions), without and with a key, an adversary with database write access edited, deleted, swapped or forged records, truncated the last 1, 10 or 100, rebuilt the chain after an edit, or restored a backup taken 50 actions earlier, at random positions in 200 trials each. Without a key, verification detects every edit, deletion and swap (200 of 200, at least 98.17\% at 95\% confidence), detects truncation, a rebuilt chain and a rollback only against an anchor, and never detects a record forged after the last anchor (Table~S9 of the supplement). With a key, it also detects every forged record and rebuilt chain, while truncation and rollback still require an anchor. As in secure audit logs \cite{schneier1999secure,crosby2009efficient}, a chain proves the order of what it holds but not its length, so the workflow manager exports an anchor at every checkpoint and workflow end. An append takes 0.06\,ms (0.29\,ms at the 95th percentile) on a ledger of one million records, so the 61 records of an average workflow add about 4\,ms, and a record takes about 2\,kB. Eight concurrent writers sustain 3{,}632 appends per second, and full verification takes 24.36\,s per million records (88.92\,s with a key).

\subsection{Energy Cost, Budgets and Review Effort}\label{sec:effort}

\begin{table}[t]
\centering
\caption{Energy per unit of work (Wh). GPU: $E_{\mathrm{tot}}$ (Eq.~\eqref{eq:energy}), with idle draw. CPU: utilization model [bound at 190\,W]. Frontier: published, inference only. Item: one alert, incident or flow answered in one request.}
\label{tab:energy}
\scriptsize
\renewcommand{\arraystretch}{0.9}
\setlength{\tabcolsep}{2pt}
\begin{tabular*}{\columnwidth}{@{\extracolsep{\fill}}l l l l@{}}
\toprule
Operation & Unit & LanG & Frontier \\
\midrule
\multicolumn{4}{@{}l}{\textit{Inference}} \\
Flow detector, batches (CPU) & flow & $4.9{\times}10^{-7}$ [$3.7{\times}10^{-6}$] & 0.24 to 0.34 \\
Flow detector, one call (CPU) & flow & $9.3{\times}10^{-5}$ [$3.6{\times}10^{-4}$] & 0.24 to 0.34 \\
Classical agents (CPU) & alert, flow & [about $10^{-5}$] & 0.24 to 0.34 \\
Alert analyst, 4B & alert & 0.0137 & 0.24 to 0.34 \\
Incident grader, 1.7B & incident & 0.0058 & 0.24 to 0.34 \\
Family namer, 8B & flow & 0.0199 & 0.24 to 0.34 \\
Untuned local models, 3B to 7B & item & 0.007 to 0.016 & 0.24 to 0.34 \\
gpt-oss-20b & item & 0.032 to 0.054 & 0.24 to 0.34 \\
Rule generator, 8 bits & attempt & 0.043 & 0.24 to 0.34 \\
\quad rule that loads and detects & rule & 0.05 to 0.06 & 0.27 to 0.39 \\
Autonomous workflow (Section~\ref{sec:evalscale}) & workflow & 0.52 & -- \\
Analysis run (Fig.~S3) & run & 22.37 & -- \\
\midrule
\multicolumn{4}{@{}l}{\textit{Training, once per model}} \\
Family classifier (CPU) & model & 2.0 [6.0] & not published \\
Site baseline fit (CPU) & tenant & $3.0{\times}10^{-4}$ [$1.5{\times}10^{-3}$] & -- \\
Site adaptation, 200 labels (CPU) & tenant & 0.12 [0.39] & -- \\
Alert analyst, 4B (all 3 steps) & adapter & 272 (837) & not published \\
Incident grader, 1.7B & adapter & 141 & not published \\
Family namer, 8B (all steps) & network & 360 to 490 (1{,}714) & not published \\
Rule generator, 1.7B (power limits) & model & at most 1{,}100 & not published \\
\bottomrule
\end{tabular*}
\end{table}

\textbf{Energy at every level.} Table~\ref{tab:energy} gathers the energy of every LanG operation and the one time cost of training each model, each costed per unit of work (a flow, an alert, an incident, a rule or a workflow), the unit a SOC processes and one frontier request would replace. Per item, the alert analyst and the incident grader use 17 to 59 times less energy than a frontier request and the family namer 12 to 17 times less, so the finetuned models that statistically match the frontier model in alert classification and rule writing use 5.6 to 59 times less energy, while untuned 3B to 7B models spend about as much for a far lower F1 (Fig.~\ref{fig:ft}) and the classical agents need no GPU. The frontier column cannot be measured from outside, and Anthropic publishes no figure for Claude Opus 5.5, so it uses published medians for one text request to comparable hosted models: 0.24\,Wh for Gemini apps, measured by Google with idle machines and data center overhead \cite{elsworth2025google}, 0.34\,Wh stated by OpenAI's CEO for ChatGPT \cite{altman2025gentle}, and about 0.3\,Wh estimated by Epoch AI for GPT-4o \cite{you2025epoch}. These cover inference only and exclude the unpublished training of the frontier model, embodied emissions and the extra energy of our long prompts \cite{you2025epoch}, so they more likely understate the frontier cost. Per rule, the range is divided by the 88.10\% of its rules that load and detect.

Training is measured, except for the rule generator, whose 8{,}907\,s run on the GPU was not metered and is bounded by the board's 320\,W limit plus 125\,W for the CPU. The adapters used 2.69\,kWh for every step of the ladder, including the steps the escalation rule discarded, or 1.28\,kg CO$_2$e at the world average intensity. Against the lower frontier estimate, the alert analyst repays the training of all its steps after 3{,}699 alerts and the incident grader after 603 incidents, about an hour of alerts for a SOC that receives 115.6 million in four years \cite{yang2024true}. On the CPU, the family classifier trains in 114\,s for 2.0\,Wh (6.0\,Wh at 190\,W), as much as scoring 4.1 million flows in batches, a site baseline is fitted in 28\,ms, and an adaptation to 200 labels takes 7.4\,s (medians of three runs), so the detector's training is negligible.

Applied to an analysis run of 22.37\,Wh, the region table gives 0.03\,g in Quebec, 1.25\,g in France and 15.9\,g in South Africa (Fig.~S3 of the supplement), a ratio above 450. A SOC running 1{,}000 such analyses a month on the world average grid emits about 10.6\,kg CO$_2$e, which a budget alert at 80\% reveals early.

Section~S12 of the supplement bounds the review effort saved analytically, which only a timed study with analysts can confirm.

\section{Discussion and Limitations}\label{sec:discussion}

Our evaluation shows that LanG's methods work under controlled conditions, not yet that they improve security operations in production. The log corpora and campaign benchmark are generated, the flows are public and the rules are replayed on traffic synthesized from each incident, which gives exact ground truth but not the noise and drift of enterprise telemetry. GUIDE and the AIT alerts are real, but GUIDE's truth is Defender's own grouping and the AIT attacks ran in simulated organizations. The frontier comparisons rest on one hosted model queried once per item, as budget and query limits allow, with unverifiable published energy estimates, samples of a few hundred items cannot resolve differences of a few points, and the carbon figures exclude embodied emissions.

The methods have limits too. Anchoring prevents false merges but cannot create missing links, and does not beat union by address on real alerts, although hub pruning keeps it accurate behind a shared egress address. On new networks, the detector holds its false alarm rate but detects 0.27\% to 92.48\% of the attacks, and only the frontier model names their families well. The rule generator still errs on YARA constraints and unfamiliar Snort and Suricata options, which the deployment checkpoint catches. Filters catch few indirect injections, so protection rests on the structural safeguards, which an adaptive attacker may probe to bias severities or summaries. The holdout gate misses poisoning that also reaches held-out labels, and the ledger needs external anchors.

Next, beachhead SOC teams will test our results on their own telemetry and time the same incidents with and without LanG, complemented by OTRF/Mordor and DARPA OpTC. The agenda follows from these limits: entity bridging from asset data, adaptive injection tests, preference training \cite{rafailov2023dpo} of the rule generator, and explanation faithfulness, while stakeholder feedback raises LanG's technology and integration readiness levels \cite{mankins1995trl,sauser2006srl}.

\section{Conclusion}\label{sec:conclusion}

We presented LanG 2.0, an agentic AI system for semi-automated security operations whose structural safeguards (entities shared on every correlation link, analyst approval of every change, a single deploying agent and a tamper-evident ledger) keep model suggestions from becoming unaudited changes. On eight flow datasets from five networks and three exporters, real Microsoft Defender XDR alerts and multistage attack campaigns, it improves on the previous LanG: requiring a shared entity prevents phase collapse and raises correlation precision from 18.75\% to 51.81\%, although grouping by address remains a strong baseline, and the 1.7B rule generator raises engine acceptance from at most 60.00\% to 98.33\%. Against Claude Opus 5.5, scoring flows against each network's normal traffic reaches 73.42\% F1 on unseen networks with no false alarm on 100 benign flows, against 58.10\% with 49.00\% false positives at no less than 650 times the energy per flow, and small finetuned models statistically match it on alert classification, incident grading and rule writing for 5.6 to 59 times less energy, while classical models match them on two of three tasks. The frontier model stays ahead only in naming attack families on a new network, until about 50 analyst labels exist. The safeguards held under prompt injection, label poisoning and ledger tampering where text filters failed, and one workstation analyzed 3.33 million records in 6.4 minutes, over 30 million an hour, without deploying a rule. Within these limits, small local models carry most SOC detection, a frontier model and a few analyst labels help on new networks, and the guarantees come from structural safeguards. With small samples and one workstation, real deployments with beachhead users come next.

\clearpage
\setcounter{section}{0}
\setcounter{figure}{0}
\setcounter{table}{0}
\setcounter{algorithm}{0}
\setcounter{equation}{0}
\renewcommand{\thetable}{S\arabic{table}}
\renewcommand{\thefigure}{S\arabic{figure}}
\renewcommand{\thealgorithm}{S\arabic{algorithm}}
\renewcommand{\thesection}{S\arabic{section}}
\renewcommand{\theequation}{S\arabic{equation}}
\twocolumn[{\centering
{\LARGE Supplementary Material for ``LanG 2.0: Energy-Efficient Agentic AI for Semi-Automated Security Operations at Scale''\par}
\vspace{4mm}
{\large Anes~Abdennebi, Nadjia~Kara, Laaziz~Lahlou and~Hakima~Ould-Slimane\par}
\vspace{6mm}}]
This supplement follows the order of the paper. Sections~\ref{S:sec:arch} to \ref{S:sec:scaling} detail the method (architecture, incident records, algorithms, carbon measurement and scalability), and Sections~\ref{S:sec:autosub} to \ref{S:sec:effort} the evaluation (autonomous subsystems, other detectors, rule generation, finetuned agents, security, targeted improvements and review effort), and Section~\ref{S:sec:avail} lists the artifacts we release. References without the prefix S point to the paper.

\section{Architecture}\label{S:sec:arch}
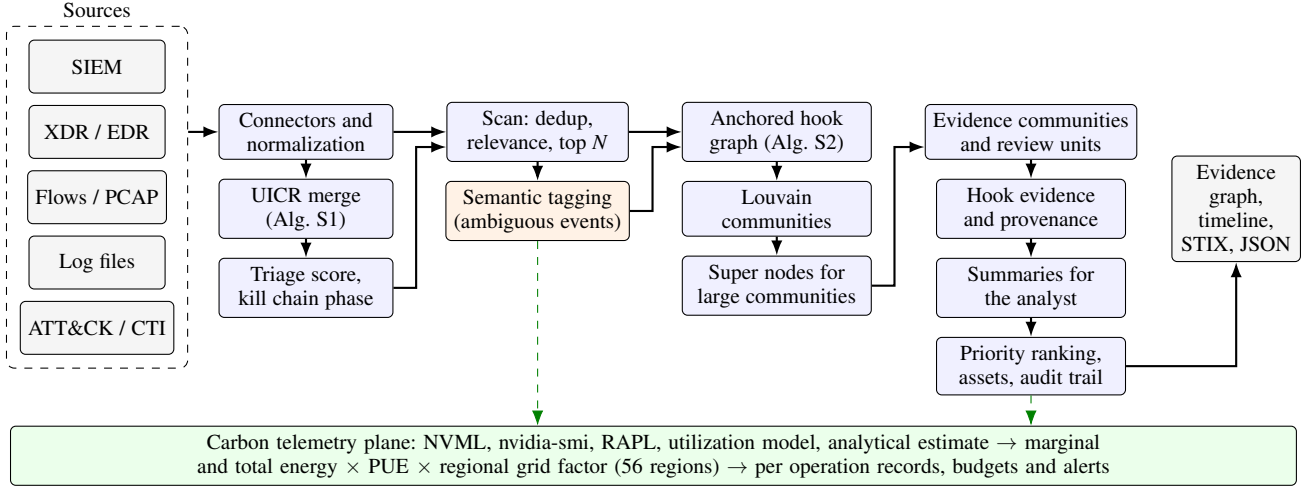
\begin{figure*}[t]
\centering
\begin{tikzpicture}[node distance=4mm and 5mm]
\node[src, minimum width=18mm] (siem) {SIEM};
\node[src, minimum width=18mm, below=1.5mm of siem] (xdr) {XDR / EDR};
\node[src, minimum width=18mm, below=1.5mm of xdr] (net) {Flows / PCAP};
\node[src, minimum width=18mm, below=1.5mm of net] (logs) {Log files};
\node[src, minimum width=18mm, below=1.5mm of logs] (cti) {ATT\&CK / CTI};
\node[grp, fit=(siem)(cti), label={[font=\footnotesize]above:Sources}] (gsrc) {};

\node[core, right=7mm of xdr, minimum width=23mm] (norm) {Connectors and\\normalization};
\node[core, below=2.5mm of norm, minimum width=23mm] (uicr) {UICR merge\\(Alg.~\ref{S:alg:uicr})};
\node[core, below=2.5mm of uicr, minimum width=23mm] (triage) {Triage score,\\kill chain phase};

\node[core, right=7mm of norm, minimum width=24mm] (scan) {Scan: dedup,\\relevance, top $N$};
\node[llm, below=2.5mm of scan, minimum width=24mm] (tag) {Semantic tagging\\(ambiguous events)};

\node[core, right=7mm of scan, minimum width=25mm] (graph) {Anchored hook\\graph (Alg.~\ref{S:alg:graph})};
\node[core, below=2.5mm of graph, minimum width=25mm] (louv) {Louvain\\communities};
\node[core, below=2.5mm of louv, minimum width=25mm] (super) {Super nodes for\\large communities};

\node[core, right=7mm of graph, minimum width=25mm] (hyp) {Evidence communities\\and review units};
\node[core, below=2.5mm of hyp, minimum width=25mm] (ground) {Hook evidence\\and provenance};
\node[core, below=2.5mm of ground, minimum width=25mm] (explain) {Summaries for\\the analyst};
\node[core, below=2.5mm of explain, minimum width=25mm] (bayes) {Priority ranking,\\assets, audit trail};

\node[blk, right=6mm of ground, minimum width=17mm, fill=gray!8] (out) {Evidence\\graph,\\timeline,\\STIX, JSON};

\draw[arr] (gsrc.east |- norm) -- (norm);
\draw[arr] (norm) -- (uicr);
\draw[arr] (uicr) -- (triage);
\draw[arr] (norm) -- (scan);
\draw[arr] (triage.east) -- ++(3mm,0) |- ([yshift=-2mm]scan.west);
\draw[arr] (scan) -- (tag);
\draw[arr] (scan) -- (graph);
\draw[arr] (tag.east) -- ++(3mm,0) |- ([yshift=-2mm]graph.west);
\draw[arr] (graph) -- (louv);
\draw[arr] (louv) -- (super);
\draw[arr] (super.east) -- ++(3mm,0) |- ([yshift=-2mm]hyp.west);
\draw[arr] (hyp) -- (ground);
\draw[arr] (ground) -- (explain);
\draw[arr] (explain) -- (bayes);
\draw[arr] (bayes.east) -| (out.south);

\node[eco, anchor=north, text width=168mm] (eco) at ([yshift=-4mm]current bounding box.center |- bayes.south) {Carbon telemetry plane: NVML, nvidia-smi, RAPL, utilization model, analytical estimate \(\rightarrow\) marginal and total energy \(\times\) PUE \(\times\) regional grid factor (56 regions) \(\rightarrow\) per operation records, budgets and alerts};
\draw[darr, green!50!black] (tag.south) -- (tag.south |- eco.north);
\draw[darr, green!50!black] (bayes.south) -- (bayes.south |- eco.north);
\end{tikzpicture}
\caption{Architecture of LanG. Orange blocks invoke a language model, metered by the carbon plane (green), and blue blocks are deterministic.}
\label{S:fig:arch}
\end{figure*}

Fig.~\ref{S:fig:arch} shows the data and evidence plane of LanG, from connectors to the evidence the agents receive.

\section{Unified Incident Context Records}\label{S:sec:uicr}
Each tool reports its own fragment of an intrusion, such as an authentication burst in the SIEM, a process tree in the EDR or a flow at a network sensor. The first stage joins these fragments into incidents with exact merge rules, so that later stages reason over incidents rather than over isolated alerts.

\subsection{The Unified Incident Context Record}

Each connector emits partial UICRs. A UICR $r$ contains an identifier, a set of correlation keys $\mathit{Keys}(r)$ such as five tuples or alert identifiers, indicators of compromise with type, value, source, first and last sighting and confidence, indicators of attack with technique, tactic and evidence, flow metadata, parsed log entries, correlated alerts, and threat features produced by local detection models. Indicators carry a fingerprint $\mathit{fp}(i)$$=H(\mathrm{type}(i)\,\Vert\,\mathrm{value}(i))$ truncated to 64 bits, which makes deduplication independent of the source that reported the indicator. The record also stores analyst metadata, namely triage score, severity, status and kill chain phase.

\subsection{Merge Rules}

Two UICRs $a$ and $b$ are merged when one of three predicates holds:
\begin{align}
P_1(a,b)&: \mathit{Keys}(a)\cap \mathit{Keys}(b)\neq\emptyset,\\
P_2(a,b)&: \{\mathit{fp}(i)\}_{i\in I(a)}\cap\{\mathit{fp}(i)\}_{i\in I(b)}\neq\emptyset,\\
P_3(a,b)&: \mathit{IP}(a)\cap\mathit{IP}(b)\neq\emptyset\wedge|\max T(a)-\min T(b)|\leq\Delta,
\end{align}
where $\mathit{IP}(\cdot)$ collects addresses from indicators and flows, $T(\cdot)$ collects log and alert timestamps and $\Delta$ is the incident window. Algorithm~\ref{S:alg:uicr} applies these rules incrementally, so a new batch never reprocesses earlier incidents unless it matches them.

\subsection{Triage Score}

After each merge, LanG recomputes a bounded triage score ($\mathrm{TS}$) in $[0,100]$:
\begin{equation}\label{S:eq:triage}
\begin{split}
\mathrm{TS}(x)=\min\Big(100,\ &\min(25,\,3|I|_{\le 5}+10\bar{c}_I)\\
&+\min(20,\,5|A|_{\le 3}+5\bar{c}_A)\\
&+\min(25,\,2|\mathcal{A}|_{\le 5}+w_{\max})\\
&+2.14\,\mathrm{ord}(\mathit{kc})+15\,m\Big),
\end{split}
\end{equation}
where $|\cdot|_{\le k}$ is a count capped at $k$, $\bar{c}_I$ and $\bar{c}_A$ are mean confidences, $\mathcal{A}$ is the alert set, $w_{\max}\in\{1,2,4,7,10\}$ is the weight of the most severe alert, $\mathrm{ord}(\mathit{kc})\in\{0,\dots,7\}$ is the rank of the furthest kill chain phase and $m$ is the highest confidence of a local model that returned a non benign label. ATT\&CK tactics \cite{S:strom2018attack} are mapped to the seven Lockheed Martin phases \cite{S:hutchins2011killchain}. Severity follows fixed bands at 20, 40, 60 and 80. The score orders the analyst queue and feeds the relevance ranking of the next stage.

\section{Algorithms}
\begin{algorithm}[!htbp]
\caption{Incremental UICR incident correlation}
\label{S:alg:uicr}
\footnotesize
\begin{algorithmic}[1]
\Require batch $B$ of partial UICRs, incident store $\mathcal{I}$, window $\Delta$
\Ensure updated incident store $\mathcal{I}$
\For{each $r\in B$}
  \State $\mathit{merged}\gets\mathit{false}$
  \For{each $x\in\mathcal{I}$}
    \If{$P_1(x,r)\vee P_2(x,r)\vee P_3(x,r)$}
      \State $x\gets x\uplus r$ \Comment{union of IoCs by $\mathit{fp}$, IoAs by (technique, tactic)}
      \State \Call{Recompute}{$x$}, $\mathit{merged}\gets\mathit{true}$, \textbf{break}
    \EndIf
  \EndFor
  \If{\textbf{not} $\mathit{merged}$}
    \State \Call{Recompute}{$r$}, $\mathcal{I}\gets\mathcal{I}\cup\{r\}$
  \EndIf
\EndFor
\State \textbf{return} $\mathcal{I}$
\Statex
\Function{Recompute}{$x$}
  \State $\mathit{kc}(x)\gets\arg\max_{a\in A(x)}\mathrm{ord}(\mathrm{KC}(\mathrm{tactic}(a)))$
  \State $\mathit{score}(x)\gets$ Eq.~\eqref{S:eq:triage}, $\sigma(x)\gets$ severity band of $\mathit{score}(x)$
\EndFunction
\end{algorithmic}
\end{algorithm}

\begin{algorithm}[!htbp]
\caption{Anchored multihook graph correlation}
\label{S:alg:graph}
\footnotesize
\begin{algorithmic}[1]
\Require primary events $E_N$, hooks $\mathcal{H}=\mathcal{H}_E\cup\mathcal{H}_C$, threshold $\theta$, damping $\gamma$
\Ensure communities $\mathcal{C}$
\State $G\gets(E_N,\emptyset)$
\For{$i\gets1$ \textbf{to} $|E_N|$}
  \For{$j\gets i+1$ \textbf{to} $|E_N|$}
    \State $\Omega\gets\{(k,h_k(\ev_i,\ev_j)) : h_k\in\mathcal{H},\ h_k(\ev_i,\ev_j)>0\}$
    \If{$\Omega$ has no entry with $k\in\mathcal{H}_E$} \textbf{continue} \EndIf
    \State $w\gets$ Eq.~\eqref{eq:composite} of the paper over $\Omega$
    \If{$w\geq\theta$} add edge $(\ev_i,\ev_j)$ with weight $w$ and evidence $\Omega$ \EndIf
  \EndFor
\EndFor
\State $\mathcal{C}\gets\textsc{Louvain}(G)$, or connected components on failure
\State remove communities with $|C|<2$
\For{each $C\in\mathcal{C}$ with $|C|>100$} fold per phase into super nodes \EndFor
\State \textbf{return} $\mathcal{C}$ sorted by size
\end{algorithmic}
\end{algorithm}

\begin{algorithm}[!htbp]
\caption{Autonomous subsystem over a large input}
\label{S:alg:autonomy}
\footnotesize
\begin{algorithmic}[1]
\Require sources $S$, detector $D$ with the tenant baseline, window $\Delta$, budget $N_u$, workers $N_w$, analyst $a$
\Ensure rule buckets $\mathcal{K}$, deployed only after analyst approval
\State $U\gets\emptyset$
\For{each chunk $X$ streamed from $S$}
  \State $U\gets U\cup\{x\in X : D \text{ flags } x\text{, or } x \text{ is severe or indicative}\}$
\EndFor
\State join $U$ into incidents with a disjoint set structure over the keys of $P_1$ and $P_2$ and over each address in buckets $\lfloor t/\Delta\rfloor$ and $\lfloor t/\Delta\rfloor-1$
\State group $U$ into work units by signature, score each unit by Eq.~\eqref{S:eq:triage} and rank the units
\State $\Pi$$\gets$ policy principal naming this subsystem and $a$
\For{each of the $N_u$ highest ranked units $w$, on $N_w$ parallel workers}
  \State run the agent graph on the evidence of $w$, answering each checkpoint as $\Pi$
  \State the deployment agent adds the rules of $w$ and their engine verdicts to $\mathcal{K}[\mathrm{family}(w)]$
\EndFor
\State \textbf{return} $\mathcal{K}$
\end{algorithmic}
\end{algorithm}

\begin{algorithm}[!htbp]
\caption{Metered model call with budget control}
\label{S:alg:carbon}
\footnotesize
\begin{algorithmic}[1]
\Require prompt $\psi$, model $\mu$, operation $o$, tenant $\ell$, region $g$, PUE, budget $(B_\ell,\alpha_\ell)$
\Ensure response $y$ and emission record $z$
\State start stopwatch, sample $P_{\mathrm{idle}}$, start probe $\mathcal{P}$ (best available method)
\State $y\gets\mu(\psi)$, collect $T_{\mathrm{in}},T_{\mathrm{out}}$ from backend metadata, else heuristic
\State stop probe, $T\gets$ elapsed
\If{$\mathcal{P}$ returned hardware energy $E_{\mathrm{tot}}$}
  \State $E_{\mathrm{marg}}\gets E_{\mathrm{tot}}-P_{\mathrm{idle}}T$, record ratio $E_{\mathrm{marg}}/E_{\mathrm{est}}$ for calibration
\Else
  \State $E_{\mathrm{tot}}\gets E_{\mathrm{marg}}\gets$ Eq.~\eqref{S:eq:analytic} $\times$ calibration factor
\EndIf
\State $z\gets(\ell,\mu,o,T,T_{\mathrm{in}},T_{\mathrm{out}},E_{\mathrm{tot}},E_{\mathrm{marg}},E\cdot\mathrm{PUE}\cdot\kappa_g,\text{method},\text{confidence})$
\State enqueue $z$ for the background writer
\If{$\mathrm{MTD}_\ell+\mathrm{CO_2e}(z)\geq\alpha_\ell B_\ell$} raise budget alert for $\ell$ \EndIf
\State \textbf{return} $y$
\end{algorithmic}
\end{algorithm}

\section{Carbon Measurement and Emissions by Region}
When no hardware counter can be read, the tracker estimates compute energy from two floating point operations per parameter per token \cite{S:kaplan2020scaling}:
\begin{equation}\label{S:eq:analytic}
E_{\mathrm{est}}=\min\Big(P_{\mathrm{peak}}T,\ \frac{2N_p(\eta T_{\mathrm{in}}+T_{\mathrm{out}})}{\Phi\,\upsilon}+P_{\mathrm{idle}}T\Big),
\end{equation}
where $N_p$ is the parameter count inferred from the model name, $T_{\mathrm{in}}$ and $T_{\mathrm{out}}$ are prompt and completion tokens, $\eta=0.1$ discounts batched prefill, $\Phi$ is the achieved FLOPs per watt of the device class, $\upsilon=0.35$ is utilization and $T$ is duration. Where hardware readings exist, the median of at least 20 reading to estimate ratios, clipped to $[0.1,10]$, calibrates estimates for that host class. Without RAPL, the CPU utilization model applies an idle floor at 30\% of the nominal package power.

The region table covers all Canadian provinces, several US states and grid operators, most EU members, major grids of other continents, three public cloud regions, a world average of 475 and a renewable contract entry at zero, from 1.2\,gCO$_2$e/kWh in Manitoba to 720 in West Virginia. A live provider can replace it. The dashboard aggregates records by operation, model, backend and measurement method, and every agent has its own operation label, which makes the most expensive step visible at a glance. The platform warns once the month to date total passes the alert level of the tenant budget, and writes happen on a background thread.
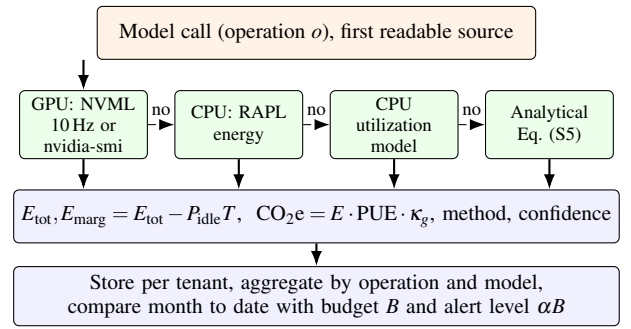
\begin{figure}[!htbp]
\centering
\begin{tikzpicture}[node distance=3mm, src4/.style={eco, font=\scriptsize, text width=15.5mm, minimum height=9mm, inner sep=2pt}]
\node[src4] (nv) {GPU: NVML 10\,Hz or nvidia-smi};
\node[src4, right=3.5mm of nv] (rapl) {CPU: RAPL energy};
\node[src4, right=3.5mm of rapl] (tdp) {CPU utilization model};
\node[src4, right=3.5mm of tdp] (est) {Analytical Eq.~\eqref{S:eq:analytic}};
\node[llm, above=4mm of $(nv.north)!0.5!(est.north)$, minimum width=58mm] (call) {Model call (operation $o$), first readable source};
\draw[arr] (call.south -| nv) -- (nv);
\draw[darr] (nv) -- node[above, font=\scriptsize]{no} (rapl);
\draw[darr] (rapl) -- node[above, font=\scriptsize]{no} (tdp);
\draw[darr] (tdp) -- node[above, font=\scriptsize]{no} (est);
\node[core, below=4mm of $(nv.south)!0.5!(est.south)$, text width=78mm] (rec) {$E_{\mathrm{tot}}, E_{\mathrm{marg}}=E_{\mathrm{tot}}-P_{\mathrm{idle}}T$, \ $\mathrm{CO_2e}=E\cdot\mathrm{PUE}\cdot{\kappa_g}$, method, confidence};
\node[core, below=3mm of rec, text width=78mm] (bud) {Store per tenant, aggregate by operation and model, compare month to date with budget $B$ and alert level $\alpha B$};
\foreach \n in {nv,rapl,tdp,est} {\draw[arr] (\n.south) -- (\n.south |- rec.north);}
\draw[arr] (rec) -- (bud);
\end{tikzpicture}
\caption{Measurement hierarchy of the carbon plane and derivation of an emission record.}
\label{S:fig:carbon}
\end{figure}

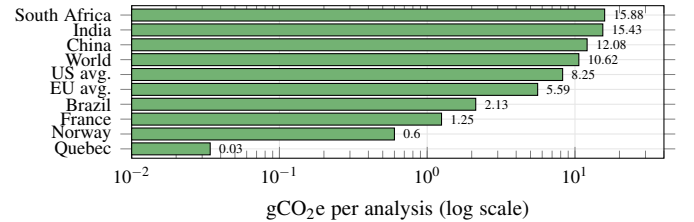
\begin{figure}[!htbp]
\centering
\begin{tikzpicture}
\begin{axis}[
  width=0.97\columnwidth, height=3.6cm,
  xbar, bar width=4.5pt, xmode=log, log origin=infty, point meta=rawx,
  xmin=0.01, xmax=40,
  symbolic y coords={Quebec,Norway,France,Brazil,EU avg.,US avg.,World,China,India,South Africa},
  ytick=data, xlabel={gCO$_2$e per analysis (log scale)},
  tick label style={font=\scriptsize}, label style={font=\footnotesize},
  nodes near coords, nodes near coords align={horizontal}, nodes near coords style={font=\tiny, /pgf/number format/fixed, /pgf/number format/precision=2},
  grid=major, grid style={gray!20}, enlarge y limits=0.07,
]
\addplot[fill=green!45!black!55] coordinates {
 (0.034,Quebec) (0.60,Norway) (1.25,France) (2.13,Brazil) (5.59,EU avg.) (8.25,US avg.) (10.62,World) (12.08,China) (15.43,India) (15.88,South Africa)};
\end{axis}
\end{tikzpicture}
\caption{Emissions of the same analysis run (22.37\,Wh total) under the grid factors of ten regions from the LanG table.}
\label{S:fig:regions}
\end{figure}

\section{Scalability of the Hook Graph}\label{S:sec:scaling}
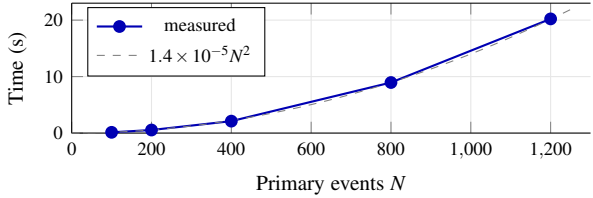
\begin{figure}[!htbp]
\centering
\begin{tikzpicture}
\begin{axis}[
  width=0.95\columnwidth, height=3.3cm,
  xlabel={Primary events $N$}, ylabel={Time (s)},
  xmin=0, xmax=1300, ymin=0, ymax=23,
  grid=major, grid style={gray!20},
  legend style={font=\scriptsize, at={(0.03,0.97)}, anchor=north west},
  tick label style={font=\scriptsize}, label style={font=\footnotesize},
  axis y line*=left,
]
\addplot[mark=*, thick, blue!70!black] coordinates {(100,0.133) (200,0.542) (400,2.115) (800,8.952) (1200,20.217)};
\addlegendentry{measured}
\addplot[dashed, gray, domain=0:1250, samples=40] {1.40e-5*x^2};
\addlegendentry{$1.4\times10^{-5}N^2$}
\end{axis}
\end{tikzpicture}
\caption{Graph construction and community detection time versus the size of the primary set (six corpora, no model calls).}
\label{S:fig:scaling}
\end{figure}

Fig.~\ref{S:fig:scaling} shows the time to evaluate all hooks, build the graph and detect communities as the primary set grows from 100 to 1{,}200 events drawn from all six corpora, without model calls. Anchoring filters edges after the hooks are computed, so it does not change this cost. The curve follows the expected quadratic trend: 0.13\,s at 100 events, 2.1\,s at 400 and 20.2\,s at 1{,}200, where the unanchored graph holds 334{,}389 edges. With the default $N=100$, this stage costs a fraction of a second whatever the raw volume.

\FloatBarrier
\section{Autonomous Subsystems}\label{S:sec:autosub}
\begin{figure}[!htbp]
\centering
\begin{tikzpicture}[x=1mm, y=1mm, font=\scriptsize,
  st/.style={draw, rounded corners=1.5pt, align=center, font=\scriptsize, minimum height=6mm,
             inner sep=1.6pt, text=black},
  cnt/.style={anchor=west, align=left, font=\scriptsize\itshape, text=black, inner sep=0.5pt},
  bucket/.style={trapezium, trapezium angle=80, trapezium stretches=true, shape border rotate=180, draw=green!45!black,
                 fill=green!10, minimum width=8mm, minimum height=7mm, inner sep=1.2pt,
                 align=center, font=\scriptsize, text=black},
  a/.style={-{Latex[length=1.8mm]}, thick, black!70}]
\node[st, fill=purple!5, draw=purple!65!black, minimum width=58mm] (launch) at (29,87.5)
  {Analysts launch autonomous subsystems\\(sources, budget {$N_u$}, workers {$N_w$})};
\node[st, fill=blue!6, minimum width=58mm] (rec) at (29,74) {Records streamed in chunks};
\node[st, fill=blue!6, minimum width=48mm] (sus) at (29,66.5) {Suspects: detector or severity};
\node[st, fill=blue!6, minimum width=38mm] (inc) at (29,59) {Incidents: $P_1$ to $P_3$};
\node[st, fill=blue!6, minimum width=28mm] (unit) at (29,51.5) {Work units, ranked};
\node[cnt] at (60,74) {2{,}365{,}424 flows\\965{,}067 log lines};
\node[cnt] at (60,66.5) {143{,}073 and\\57{,}041 suspects};
\node[cnt] at (60,59) {6 and 57{,}041};
\node[cnt] at (60,51.5) {238 and 577};
\node[draw, dashed, rounded corners=2pt, inner sep=2.2pt, fit=(rec)(unit)(sus)] (front) {};
\node[anchor=south west, font=\scriptsize, inner sep=0.5pt, text=black] at (front.north west) {front stage, linear time};
\node[st, fill=orange!6, minimum width=56mm, minimum height=11mm] at (31.6,36.6) {};
\node[st, fill=orange!8, minimum width=56mm, minimum height=11mm] at (30.3,35.3) {};
\node[st, fill=orange!10, minimum width=56mm, minimum height=11mm] (sub) at (29,34)
  {Subsystem $k$: agent graph of Fig.~\ref{fig:agentic} of the paper\\on the {$N_u$} top units, {$N_w$} in parallel,\\checkpoints answered by policy {$\Pi_k$}};
\node[cnt] at (60,34) {2 subsystems,\\25 workflows each};
\node[bucket, anchor=west] (b1) at (-1,20) {DoS};
\node[bucket, right=0.9mm of b1] (b2) {recon};
\node[bucket, right=0.9mm of b2] (b3) {exploit};
\node[bucket, right=0.9mm of b3] (b4) {brute\\force};
\node[bucket, right=0.9mm of b4] (b5) {malware\\and C2};
\node[bucket, right=0.9mm of b5] (b6) {collection};
\node[bucket, right=0.9mm of b6] (b7) {logs, by\\service};
\coordinate (bustop) at (0,26.3);
\coordinate (busbot) at (0,14.3);
\node[cnt] at (58,8.5) {207 rules in six buckets,\\none deployed};
\node[st, fill=purple!5, draw=purple!65!black, dashed, minimum width=50mm] (rev) at (29,8.5) {Analysts or SOC team review buckets};
\node[st, fill=blue!6, minimum width=26mm] (dep) at (13,-4.5) {Deployment agent\\writes rule};
\node[st, fill=gray!8, minimum width=26mm] (rej) at (45,-4.5) {Rule closed,\\not deployed};
\draw[a] (launch) -- (front);
\draw[a] (front) -- (sub);
\draw[thick, black!70] (sub.south) -- (sub.south |- bustop) (b1.north |- bustop) -- (b7.north |- bustop);
\draw[thick, black!70] (b1.south |- busbot) -- (b7.south |- busbot);
\foreach \b in {b1,b2,b3,b4,b5,b6,b7} {
  \draw[a] (\b.north |- bustop) -- (\b.north);
  \draw[thick, black!70] (\b.south) -- (\b.south |- busbot);
}
\draw[a] (rev.north |- busbot) -- (rev.north);
\draw[a] (rev.south) -- ++(0,-2.5) -| node[pos=0.75, left, font=\scriptsize, text=black] {approve} (dep.north);
\draw[a] (rev.south) -- ++(0,-2.5) -| node[pos=0.75, right, font=\scriptsize, text=black] {reject} (rej.north);
\end{tikzpicture}
\caption{Autonomous subsystems. A front stage turns millions of records into ranked work units, each subsystem runs the agent graph (Fig.~\ref{fig:agentic} of the paper) on its top units with a policy at the checkpoints, and rules wait in buckets for an analyst. Counts from Section~\ref{sec:evalscale} of the paper.}
\label{S:fig:autonomy}
\end{figure}
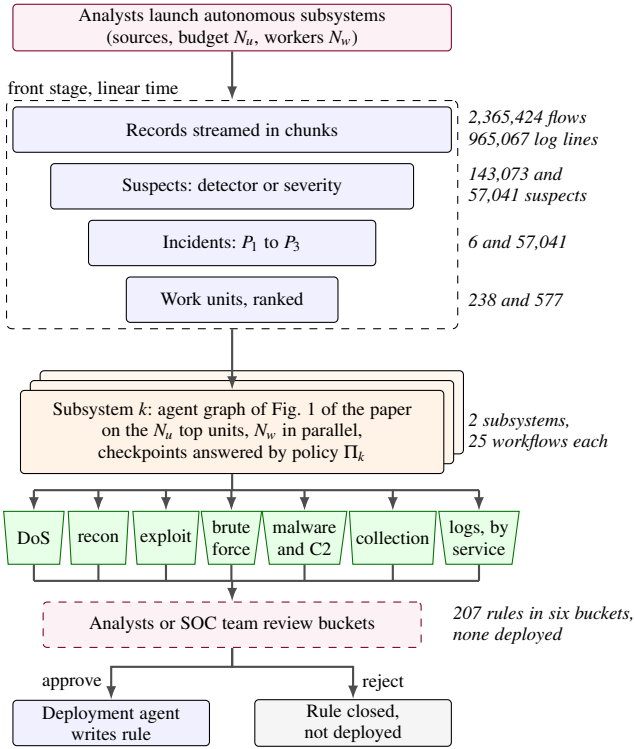

\begin{table}[!htbp]
\centering
\caption{Two autonomous subsystems on 3.33 million records. Attack shares use labels the subsystems never read.}
\label{S:tab:autonomy}
\scriptsize
\setlength{\tabcolsep}{3pt}
\begin{tabular*}{\columnwidth}{@{\extracolsep{\fill}}l r r@{}}
\toprule
 & Flows & Logs \\
\midrule
Records & 2{,}365{,}424 & 965{,}067 \\
Suspects after the filter & 143{,}073 & 57{,}041 \\
Attack records kept by the filter & 92.69\% & 68.37\% \\
Incidents, work units & 6, 238 & 57{,}041, 577 \\
Front stage time, throughput & 51.5\,s, 45{,}957/s & 18.2\,s, 52{,}822/s \\
Workflows run (failed), policy decisions & 25 (0), 50 & 25 (0), 50 \\
Workflow time, median and 95th percentile & 48.1\,s, 74.1\,s & 48.4\,s, 67.6\,s \\
Total time & 382.2\,s & 322.7\,s \\
Attack suspects in the units analyzed & 93.17\% & 97.20\% \\
Attack classes or techniques covered & 9 of 9 & 2 of 2 \\
Rules in buckets, accepted by engines & 100, 98 & 107, 103 \\
Rules deployed without approval & 0 & 0 \\
\bottomrule
\end{tabular*}
\end{table}

The flows formed only six incidents, because a few addresses of UNSW-NB15 chain most suspects through the address rule, whereas the generated logs give each attack event its own address, so no two log suspects merged, and the work units carried the reduction. One workflow per suspect would take 337.5 hours at the measured mean workflow times and eight workflows at a time. The emission records of the individual calls sum to 141.6\,Wh against 25.99\,Wh measured for the devices, of which 23.09\,Wh on the GPU. Our first ranking, by severity band and then by size, analyzed units that held only 0.52\% of the attack log lines and 63.95\% of the attack flows, as error lines of benign timeouts outranked warning lines of brute force. Weighting the triage score by the logarithm of size corrected this without labels.

\section{Other Detection Models}\label{S:sec:otherdet}
We also retrained the text classifier used in our previous work \cite{S:abdennebi2026lang}, Prompt Guard~2 (86M parameters) \cite{S:chennabasappa2025llamafirewall}, with LoRA on the same eight datasets. On identical held out time blocks, boosted trees reach a family macro F1 of 82.92\% against 81.21\%, win on all eight datasets, and take a median of 0.86\,ms per flow on the CPU against 12.16\,ms on the GPU. Finally, the model for DNS over HTTPS used in our previous work \cite{S:abdennebi2026lang} (weighted F1 of 99.00\%) is replaced by boosted trees on BCCC-CIRA-CIC-DoHBrw-2020 \cite{S:montazeri2020doh}, with an F1 of 99.99\% after removing 1.43\% duplicate rows, in 0.49\,ms per flow.

\section{Rule Generation}\label{S:sec:rules}
\begin{table}[!htbp]
\centering
\caption{Rule generation on 240 held out incidents (\%), the values of Fig.~\ref{fig:rules} of the paper. Acc.: rules the engine loads. Det.: the 84 replayable incidents detected. Recall: given indicators used. ``+ loop'': validator loop, at most three attempts.}
\label{S:tab:rules}
\scriptsize
\setlength{\tabcolsep}{2.2pt}
\begin{tabular*}{\columnwidth}{@{\extracolsep{\fill}}l ccccc@{}}
\toprule
System & Acc. & Det. & Recall & Wh per rule & p50 (s) \\
\midrule
\multicolumn{6}{@{}l}{\textit{Local models finetuned for rule generation (previous work in 4 bits)}} \\
LanG v2 + loop, 16 bits & \textbf{98.33} & \textbf{82.14} & 98.41 & 0.128 & 3.37$^\dagger$ \\
LanG v2 + loop, 8 bits & 97.50 & 79.76 & 98.37 & \textbf{0.047} & 1.00$^\dagger$ \\
LanG v2, 8 bits, one attempt & 95.00 & 82.14 & 98.41 & \textbf{0.043} & \textbf{0.78} \\
Previous work \cite{S:abdennebi2026lang}, Qwen3 1.7B$^\ast$ & 60.00 & 29.03 & 73.86 & 0.169 & 7.00 \\
Previous work \cite{S:abdennebi2026lang}, Mistral 7B$^\ast$ & 57.50 & 29.03 & 79.25 & 0.352 & 9.81 \\
\midrule
\multicolumn{6}{@{}l}{\textit{Local models not finetuned for rule generation (Ollama, 4 bits)}} \\
Qwen 2.5 Coder 7B & 54.17 & 41.67 & 91.41 & 0.119 & 1.32 \\
Llama 3.1 8B & 31.25 & 8.33 & 86.43 & 0.109 & 1.30 \\
Llama 3.2 3B & 1.67 & 0.00 & 41.39 & 0.061 & 0.81 \\
\midrule
Claude Opus 5.5 & 99.58 & 88.10 & 98.93 & 0.24 to 0.34 & n/a \\
\bottomrule
\end{tabular*}
\par
{\raggedright $^\ast$Not trained on Sigma: 200 incidents, 62 replayable. Our previous work reported 11.2 and 13.65\,s per rule on its own prompts. $^\dagger$From the full run, with validation.}\par
\end{table}

\subsection{Complete Workflows}
In complete orchestrator runs with the new detector and the served generator (1.3\,s per rule), the deployment checkpoint was reached in both workflows that produced rules. Rules for these incidents, unlike the training data, failed their engine after three attempts through a \texttt{depth} shorter than its content, an unknown Snort~2 classtype or a mismatched Suricata~7 buffer size. The analyst withheld them, the other rules were deployed, and the ledger chain verified in every workflow.

\section{Finetuned Agents: Protocol and Ladder}\label{S:sec:ftproto}
\begin{table}[!htbp]
\centering
\caption{Macro F1 (\%) of the detection phase agents (Fig.~\ref{fig:ft} of the paper) on unseen organizations and networks. Sub.: items the frontier model answered (397, 250, 217). All: whole test set (10{,}297, 3{,}000, 5{,}495), run for the finetuned and classical systems only (Section~\ref{sec:evalft} of the paper explains the empty cells). Classical: TF-IDF with logistic regression (alerts), LightGBM (others). 95\% intervals in the text.}
\label{S:tab:ft}
\scriptsize
\setlength{\tabcolsep}{2.4pt}
\begin{tabular*}{\columnwidth}{@{\extracolsep{\fill}}l cc cc cc@{}}
\toprule
& \multicolumn{2}{c}{Alert step} & \multicolumn{2}{c}{Incident grade} & \multicolumn{2}{c}{Flow family} \\
\cmidrule(lr){2-3}\cmidrule(lr){4-5}\cmidrule(lr){6-7}
System & Sub. & All & Sub. & All & Sub. & All \\
\midrule
\textbf{LanG finetuned} & 97.31 & 90.75 & 42.83 & 40.95 & 12.14 & 16.27 \\
\quad base size & \multicolumn{2}{c}{4B} & \multicolumn{2}{c}{1.7B} & \multicolumn{2}{c}{8B} \\
Claude Opus 5.5 & 95.19 &  & 27.24 &  & 26.78 &  \\
gpt-oss-20b & 53.64 &  & 30.07 &  & 14.78 &  \\
Classical & 98.32 & 93.10 & 43.41 & 42.16 & 12.24 & 12.58 \\
Keyword rules & 56.00 & 42.86 &  &  &  &  \\
Llama 3.2 3B & 18.73 &  & 19.18 &  & 7.32 &  \\
Qwen 3.5 4B & 37.98 &  & 18.49 &  & 10.64 &  \\
Qwen 2.5 7B & 36.83 &  & 23.42 &  & 11.91 &  \\
\bottomrule
\end{tabular*}
\end{table}

\textbf{Data.} The alert analyst uses the AIT alert data set: alerts of five organizations for training, one for validation and two (10{,}297 alerts) for test, with at most five alerts per masked template in training. The incident grader uses the GUIDE training file, split by organization 70/10/20, with 12{,}000 training incidents (at most 300 per organization) and 3{,}000 test incidents at natural prevalence, without the post triage fields ActionGrouped and ActionGranular. The family namer holds out one network at a time (UNSW-NB15, ToN-IoT, CSE-CIC-IDS2018), trains on the others, and uses BoT-IoT as a development fold for the escalation rule. Its test sets hold attack flows and benign flows that the site baseline flagged, 5{,}495 distinct flows in all.

\textbf{Training and serving.} Each adapter is a LoRA of rank 16 ($\alpha=32$) on all projections, trained for two epochs at a learning rate of $2\times10^{-4}$ on the completion only, with 4 bit QLoRA for 8B bases. The adapter is merged, converted to GGUF at 8 bits and served by Ollama at temperature 0 with a JSON schema, at most 48 new tokens (64 for alerts) and a context of 4{,}096 tokens. A response that does not parse or names a class outside the task counts as wrong. The 8B adapters are merged shard by shard and served from the same 8 bit GGUF by llama.cpp, which maps the file in place, since importing 8.7\,GB models into Ollama exhausted the memory of the workstation. Decoding is identical.

\textbf{Frontier protocol.} Each frontier answer comes from a fresh Claude Opus 5.5 context that receives one line asking for the answer only, without tools, followed by the same instruction and item as the local models. Repeated item texts were asked once. The frontier model answered 397 test and 109 validation alerts, 250 test and 30 validation incidents, and 217 test and 39 development flows. Its energy is the published 0.24 to 0.34\,Wh per request. The open weight gpt-oss-20b \cite{S:openai2025gptoss} (MXFP4 GGUF) is served by llama.cpp with its chat template, low reasoning effort, greedy decoding, seed 0 and the same JSON schema, and its energy is measured on our GPU like the local models.

\begin{table}[!htbp]
\centering
\caption{Escalation ladder: validation macro F1 (\%) on items the frontier model also answered, and training energy. A larger base is tried only if the agent trails the frontier model and its last step gained at least two points.}
\label{S:tab:ladder}
\scriptsize
\begin{tabular*}{\columnwidth}{@{\extracolsep{\fill}}l l r r r r l@{}}
\toprule
Agent & Base & F1 & Frontier & Gain & kWh & Decision \\
\midrule
Alert analyst & 1.7B & 85.10 & 89.76 &  & 0.136 & escalate \\
(109 alerts) & 4B & 89.63 & 89.76 & 4.53 & 0.272 & escalate \\
 & 8B & 72.97 & 89.76 & $-16.66$ & 0.429 & keep 4B \\
\midrule
Incident grader & 1.7B & 55.13 & 29.40 &  & 0.141 & keep \\
(30 incidents) & & & & & & \\
\midrule
Family namer & 1.7B & 17.52 & 52.61 &  & 0.130 & escalate \\
(39 flows) & 4B & 20.75 & 52.61 & 3.23 & 0.314 & escalate \\
 & 8B & & & & 0.36--0.49 & per fold \\
\bottomrule
\end{tabular*}
\end{table}

\begin{table}[!htbp]
\centering
\caption{Family namer on each test network, absent from the training of both systems. Macro F1 (\%) with its 95\% confidence interval (CI) over flows.}
\label{S:tab:t1fold}
\scriptsize
\begin{tabular*}{\columnwidth}{@{\extracolsep{\fill}}l r c c@{}}
\toprule
& & \multicolumn{2}{c}{Macro F1 [95\% CI]} \\
\cmidrule(l){3-4}
Network & Flows & Finetuned 8B & LightGBM \\
\midrule
UNSW-NB15 & 2{,}535 & 9.20 [8.34, 10.11] & 11.43 [10.52, 12.42] \\
ToN-IoT & 2{,}405 & 19.20 [17.88, 20.69] & 10.06 [9.25, 10.97] \\
CSE-CIC-IDS2018 & 1{,}059 & 15.65 [14.19, 17.37] & 13.65 [12.21, 15.07] \\
\midrule
All (distinct flows) & 5{,}495 & 16.27 [15.35, 17.26] & 12.58 [11.92, 13.23] \\
\bottomrule
\end{tabular*}
\end{table}

\FloatBarrier
\section{Security Evaluation Details}

\begin{table}[!htbp]
\centering
\caption{Input filters on published injections and benign text, each passed through the production quarantine as one log message. Share flagged, and quarantined (never seen by a model), in \% with 95\% confidence interval (CI). Pattern layer: 0.1\,ms per field. Prompt Guard~2: about 10\,ms more on the GPU.}
\label{S:tab:guard}
\scriptsize
\setlength{\tabcolsep}{1.5pt}
\begin{tabular*}{\columnwidth}{@{\extracolsep{\fill}}l c c c@{}}
\toprule
& Pattern layer & \multicolumn{2}{c}{With Prompt Guard 2} \\
& \multicolumn{3}{c}{\% [95\% CI]} \\
\cmidrule(lr){2-2}\cmidrule(lr){3-4}
Texts & flagged & flagged & quarantined \\
\midrule
BIPIA instructions (150) & 0.00 [0.00, 2.43] & 0.67 [0.12, 3.68] & 0.00 [0.00, 2.43] \\
BIPIA in log lines (150) & 0.00 [0.00, 2.43] & 0.67 [0.12, 3.68] & 0.00 [0.00, 2.43] \\
deepset injections (263) & 5.32 [3.20, 8.74] & 26.24 [21.29, 31.87] & 19.39 [15.07, 24.60] \\
deepset benign (399) & 0.00 [0.00, 0.92] & 0.25 [0.04, 1.41] & 0.00 [0.00, 0.92] \\
AIT alerts (2{,}000) & 0.00 [0.00, 0.18] & 0.00 [0.00, 0.18] & 0.00 [0.00, 0.18] \\
AIT log lines (2{,}000) & 0.00 [0.00, 0.18] & 0.00 [0.00, 0.18] & 0.00 [0.00, 0.18] \\
\bottomrule
\end{tabular*}
\end{table}

\begin{table}[!htbp]
\centering
\caption{Injection in complete workflows: 40 AIT incidents (30 attacks, 10 benign), one BIPIA attack each in a user agent or log line. Changes against the same incident without attack, in \% of 40 workflows. Clean rerun: identical clean prompts run twice. Benign: attack incidents graded benign. All workflows reached the analyst.}
\label{S:tab:injsys}
\scriptsize
\setlength{\tabcolsep}{2pt}
\begin{tabular*}{\columnwidth}{@{\extracolsep{\fill}}l cccccc@{}}
\toprule
& & \multicolumn{2}{c}{Severity} & Priority & Log risk & \\
\cmidrule(lr){3-4}
Configuration & Flagged & lower & higher & lower & changed & Benign \\
\midrule
No filter & 0.00 & 15.00 & 25.00 & 0.00 & 47.50 & 0 of 30 \\
Pattern filter & 0.00 & 30.00 & 20.00 & 2.50 & 55.00 & 0 of 30 \\
Prompt Guard 2, spotlighting & 0.00 & 37.50 & 30.00 & 10.00 & 55.00 & 0 of 30 \\
Benign text control & 0.00 & 32.50 & 30.00 & 7.50 & 45.00 & 0 of 30 \\
Clean rerun & n/a & 10.00 & 32.50 & 0.00 & 57.50 & n/a \\
\bottomrule
\end{tabular*}
\end{table}

\begin{table}[!htbp]
\centering
\caption{Poisoned feedback, mean of 12 runs (four NetFlow data sets, three baseline draws), in \%. FPR and Det.: alerts on benign and attack flows. Fam.\ F1: naming of detected attacks. Named, Alerted: targeted family flows named correctly, alerted. Gate: adaptations accepted (of 12). Audit: poisoned labels sent to a second analyst. Without labels: FPR 1.08, Det.\ 59.48, Fam.\ F1 0.86, Named 2.26, Alerted 63.54.}
\label{S:tab:poison}
\scriptsize
\setlength{\tabcolsep}{2.2pt}
\begin{tabular*}{\columnwidth}{@{\extracolsep{\fill}}l ccccc cc@{}}
\toprule
Labels & FPR & Det. & Fam.\ F1 & Named & Alerted & Gate & Audit \\
\midrule
Clean & 2.57 & 89.24 & 61.01 & 80.97 & 92.86 & 9 & n/a \\
Targeted, $\epsilon=2\%$ & 2.56 & 83.57 & 59.38 & 70.12 & 84.87 & 8 & 70.83 \\
Targeted, $\epsilon=5\%$ & 2.64 & 78.53 & 56.53 & 55.95 & 75.59 & 7 & 66.11 \\
Targeted, $\epsilon=10\%$ & 2.31 & 73.00 & 53.74 & 38.34 & 68.07 & 10 & 63.89 \\
Targeted, $\epsilon=20\%$ & 1.82 & 69.00 & 43.51 & 10.97 & 64.12 & 12 & 65.00 \\
Availability, $\epsilon=2\%$ & 2.47 & 88.40 & 60.34 & 78.68 & 91.24 & 12 & 0.00 \\
Availability, $\epsilon=5\%$ & 2.64 & 89.19 & 60.47 & 79.86 & 92.09 & 11 & 0.00 \\
Availability, $\epsilon=10\%$ & 2.91 & 88.60 & 59.40 & 80.01 & 92.05 & 12 & 0.00 \\
Availability, $\epsilon=20\%$ & 7.79 & 89.04 & 61.18 & 80.52 & 92.28 & 6 & 0.00 \\
Autonomous self labels & 1.07 & 59.48 & 0.59 & 1.38 & 63.53 & 9 & 60.80 \\
\bottomrule
\end{tabular*}
\end{table}

\begin{table}[!htbp]
\centering
\caption{Ledger tampering: trials detected of 200 per attack, by the chain alone and against an anchor taken before the attack.}
\label{S:tab:tamper}
\scriptsize
\begin{tabular*}{\columnwidth}{@{\extracolsep{\fill}}l cc cc@{}}
\toprule
& \multicolumn{2}{c}{Without key} & \multicolumn{2}{c}{With HMAC key} \\
\cmidrule(lr){2-3}\cmidrule(lr){4-5}
Attack & chain & anchor & chain & anchor \\
\midrule
Edit one record & 200 & 200 & 200 & 200 \\
Delete one record & 200 & 200 & 200 & 200 \\
Truncate 1, 10 or 100 records & 0 & 200 & 0 & 200 \\
Forge a record after the head & 0 & 0 & 200 & 200 \\
Swap two records & 200 & 200 & 200 & 200 \\
Edit and rebuild the chain & 0 & 200 & 200 & 200 \\
Restore a backup & 0 & 200 & 0 & 200 \\
\bottomrule
\end{tabular*}
\end{table}

\FloatBarrier
\section{Targeted Improvements}\label{S:sec:improve}
After the main evaluation, we applied six inexpensive fixes to the weakest results, measured on the same data and splits, outside the preregistered comparisons. Table~\ref{S:tab:improve} summarizes them. All runs together took under two hours on the workstation.

\begin{table}[!htbp]
\centering
\caption{Targeted improvements, before and after (\%). F1 is macro F1 except for correlation (B$^3$ F1). Means over five draws for the label experiments.}
\label{S:tab:improve}
\scriptsize
\setlength{\tabcolsep}{2pt}
\begin{tabular*}{\columnwidth}{@{\extracolsep{\fill}}l l r r@{}}
\toprule
Fix & Metric & Before & After \\
\midrule
Greedy decoding, fixed seed & severity changed, clean rerun & 42.50 & 0.00 \\
 & log risk changed, clean rerun & 57.50 & 25.00 \\
50 site labels (family namer) & F1, three new networks & 12.31 & 55.07 \\
\quad 200 site labels & & & 67.08 \\
\quad frontier names as labels & & & 23.33 \\
10 labels per organization & F1, 17 GUIDE organizations & 28.25 & 69.53 \\
\quad 25 labels & & & 77.09 \\
Hub pruning, anchored graph & F1, Benchmark B, $\eta=1$ & 24.49 & 41.18 \\
 & F1, GUIDE, one day & 37.04 & 39.39 \\
YARA repair (validator loop) & detection, 84 incidents & 82.14 & 86.90 \\
 & detection, 108 held out YARA & 57.41 & 80.56 \\
Score ensemble (not adopted) & detection at 1\% false alarms & 42.66 & 24.38 \\
\bottomrule
\end{tabular*}
\end{table}

\FloatBarrier
\section{Review Effort}\label{S:sec:effort}
Besides energy, the workflow changes how much evidence an analyst must review. Let $t_r$ be the time to inspect one event, $n_e$ the incoming event count and $n_c$ the number of evidence units returned for review. Manual correlation requires at least $n_e t_r$, while the proposed workflow requires $n_c t_r$$+T_m$, where $T_m$ is machine time. In the measured run, $n_e=100$, $n_c=11$ and $T_m=261$ seconds. With $t_r=2$ minutes, the manual lower bound is 200 minutes and the workflow takes about 26 minutes, a reduction of 7.6 times. This is an analytical estimate under a stated assumption on review time, not a field study or evidence of a shorter time to identify breaches. Measuring that time requires repeated investigations of the same scenarios by analysts working manually and with LanG, which we plan with beachhead users (Section~\ref{sec:discussion} of the paper).

\FloatBarrier

\section{Artifact Availability}\label{S:sec:avail}
All datasets are public: the NetFlow~v3, CICFlowMeter and NTLFlowLyzer exports of Section~\ref{sec:evalnet} of the paper, GUIDE, the AIT alert data set, Emerging Threats Open, the Snort community rules, SigmaHQ, public YARA rules, BIPIA and the deepset injection collection. With the paper we release the LanG code, the benchmark and figure scripts, the generated log corpora and their generators, the identifiers and seeds of every evaluated item, the preregistration of the finetuned agents, the prompts, the frontier answers and all result files in the repository \mbox{\url{https://github.com/RoronoaZ/LanG}}. The finetuned weights are available on request.

\end{document}